\documentclass[10pt,journal,letterpaper,compsoc]{IEEEtran}

\usepackage[cmex10]{amsmath}
\usepackage{amssymb,bm,graphicx}
\usepackage{cite}
\usepackage[normalem]{ulem}
\usepackage[colorlinks=true]{hyperref}
\usepackage[T1]{fontenc}
\usepackage{rotating}

\newcommand{\etc}{{\em etc}.~}
\newcommand{\eg}{{\em eg.}}
\newcommand{\ie}{{\em i.e.}}
\newcommand{\etal}{{\em et al.}}

\DeclareMathOperator*{\argmin}{arg\,min}

\begin{document}

\title{Modeling Radiometric Uncertainty for\\Vision with Tone-mapped Color Images}
\author{%
Ayan~Chakrabarti, Ying~Xiong, Baochen~Sun, Trevor~Darrell,\\%
Daniel~Scharstein, Todd~Zickler, and Kate Saenko%
\IEEEcompsocitemizethanks{%
\IEEEcompsocthanksitem AC, YX, and TZ are with the Harvard School of Engineering and Applied Sciences, Cambridge, MA 02138. BS and KS are with the University of Massachusetts Lowell, Lowell, MA 01854. TD is with the University of California Berkeley, Berkeley, CA 094720. DS is with Middlebury College, Middlebury, VT 05753. E-mails: \{ayanc@eecs.harvard.edu, yxiong@seas.harvard.edu, bsun@cs.uml.edu, trevor@eecs.berkeley.edu, schar@middlebury.edu, zickler@seas.harvard.edu, saenko@cs.uml.edu\}.
}}

\IEEEcompsoctitleabstractindextext{%
\begin{abstract}

To produce images that are suitable for display, tone-mapping is widely used in digital cameras to map linear color measurements into narrow gamuts with limited dynamic range. This introduces non-linear distortion that must be undone, through a radiometric calibration process, before computer vision systems can analyze such photographs radiometrically. This paper considers the inherent uncertainty of undoing the effects of tone-mapping. We observe that this uncertainty varies substantially across color space, making some pixels more reliable than others. We introduce a model for this uncertainty and a method for fitting it to a given camera or imaging pipeline. Once fit, the model provides for each pixel in a tone-mapped digital photograph a probability distribution over linear scene colors that could have induced it. We demonstrate how these distributions can be useful for visual inference by incorporating them into estimation algorithms for a representative set of vision tasks.

\end{abstract}


}
\maketitle

\section{Introduction}\label{sec:introduction}

The proliferation of digital cameras has created an explosion of photographs being shared online. Most of these photographs exist in narrow-gamut, low-dynamic range formats---typically those defined in the sRGB or Adobe RGB standards---because they are intended primarily for display through devices with limited gamut and dynamic range. While this workflow is efficient for storage, transmission, and display-processing, it is unfortunate for computer vision systems that seek to exploit online photo collections to learn object appearance models for recognition; reconstruct three-dimensional (3D) scene models for virtual tourism; enhance images through processes like denoising and deblurring; and so on. Indeed, many of the computer vision algorithms required for these tasks use radiometric reasoning and therefore assume that image color values are directly 
\begin{figure}[!t]
\centering
\includegraphics[width=\columnwidth]{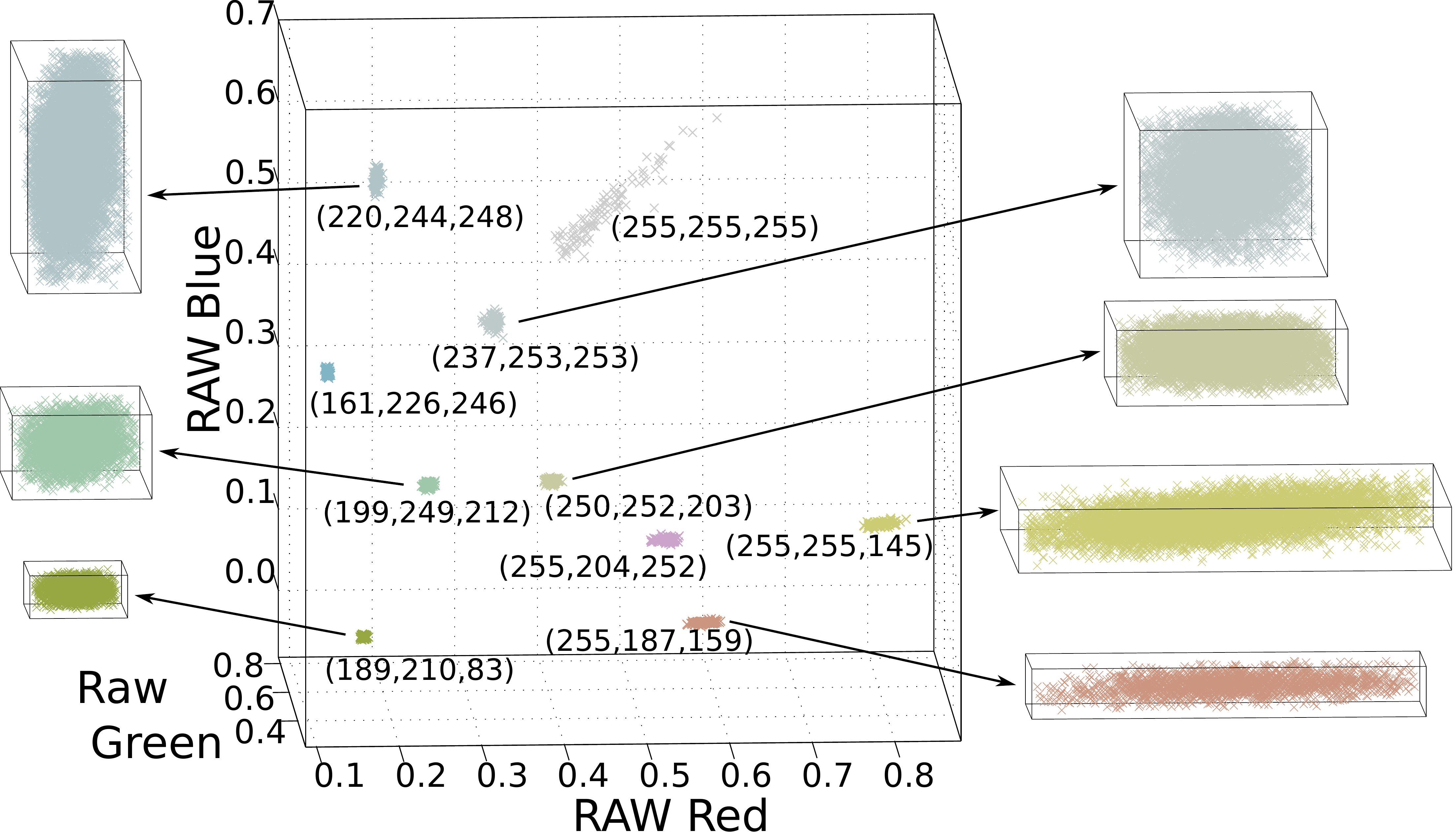}
\caption{Clusters of RAW measurements that each map to a single JPEG color value (indicated in parentheses) in a digital SLR camera (Canon EOS 40D). Close-ups of the clusters emphasize the variations in cluster size and orientation. When inverting the tone-mapping process, this structured uncertainty cannot be avoided.}\label{fig:teaser}
\end{figure}
proportional to spectral scene radiance (called \emph{RAW color} hereafter). But when a consumer camera renders---or globally tone-maps---its digital linear color measurements to an output-referred, narrow-gamut color encoding (called \emph{JPEG color} hereafter), this proportionality is almost always destroyed.\footnote{Some comments on terminology. We use colloquial phrases \emph{RAW color} and \emph{JPEG color} respectively for linear, scene-referred color and non-linear, output-referred color. The latter does not include lossy compression, and should not be confused with JPEG compression. Also, we use \emph{(global) tone-map} for any spatially-uniform, non-linear map of each pixel's color, independent of the values of its surrounding pixels. It is nearly synonymous with the common phrase ``radiometric response function''~\cite{mitsunaga1999rsc}, but generalized to include cross-channel maps.}

\begin{figure*}[t]
\centering
\includegraphics[width=\textwidth]{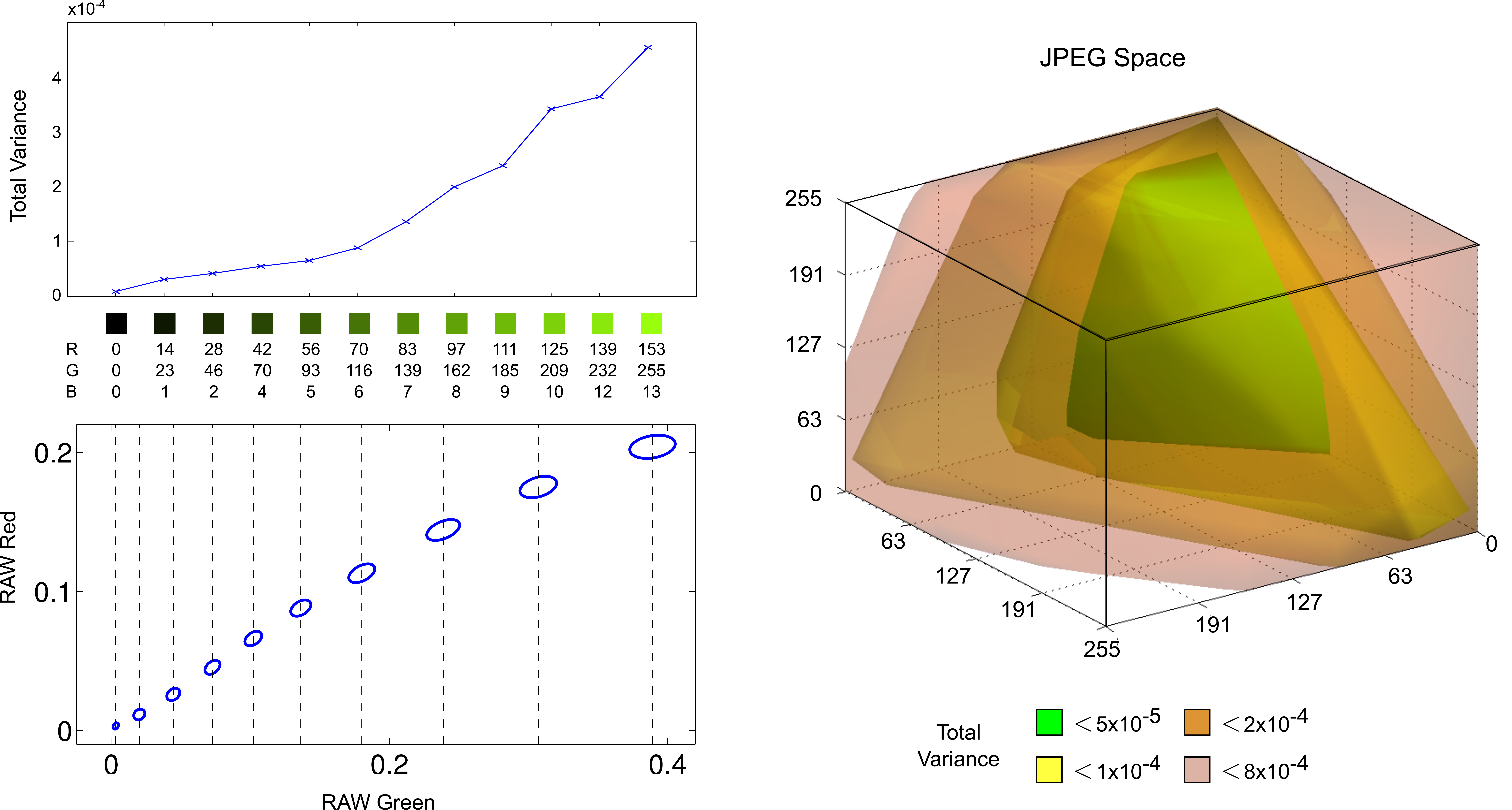}
\caption{Derendering Uncertainty (for the \emph{Canon PowerShot S90}). (Left) Total variance  of the distributions of RAW values that are tone-mapped to a set of rendered color values that lie along a line in JPEG space; along with ellipses depicting corresponding distributions along two dimensions in the RAW sensor space. (Right) 3D volumes depicting the change in derendering variance across JPEG space.}
\label{fig:ResponseFunc} 
\end{figure*}

In computer vision, we try to undo the non-linear effects of tone-mapping so that radiometric reasoning about consumer photographs can be  more effective. To this end, there are many methods for fitting parametric forms to the global tone-mapping operators applied by color cameras---so-called ``radiometric calibration" methods~\cite{debevec97hdr, mann1995bud, mitsunaga1999rsc, emor, bmvc09, kim2012new, lee2013radiometric}---and it is now possible to fit many global tone-mapping operators with high precision and accuracy~\cite{kim2012new}. However, once these maps are estimated, standard practice for undoing color distortion in observed non-linear JPEG colors is to apply a simple inverse mapping in a one-to-one manner~\cite{debevec97hdr, mann1995bud, mitsunaga1999rsc, emor, bmvc09, kim2012new, lee2013radiometric}. This ignores the critical fact that forward tone-mapping processes lead to loss of information that is highly structured.

Tone-mapping is effective when it leads to narrow-gamut images that are nonetheless visually-pleasing, and this necessarily involves non-linear compression. Once the compressed colors are quantized, the reverse mapping becomes one-to-many as shown in Fig.~\ref{fig:teaser}, with each nonlinear JPEG color being associated with a \emph{distribution} of linear RAW colors that can induce it. The amount of color compression in the forward tone-map, as well as the (hue/lightness) directions in which it occurs, change considerably across color space. As a result, the variances of reverse-mapped RAW color distributions unavoidably span a substantial range, with some predicted linear RAW colors being much more reliable than others.

How can we know which predicted RAW colors are unreliable? Intuitively, the forward compression (and thus the reverse uncertainty) should be greatest near the boundary of the output gamut, and practitioners often leverage this intuition by heuristically ignoring all JPEG pixels that have values above or below certain thresholds in one or more of their channels. However, as shown in Fig.~\ref{fig:ResponseFunc}, the variances of inverse RAW distributions tend to change \emph{continuously} across color space, and this makes the choice of such thresholds arbitrary. Moreover, this heuristic approach relies on discarding information that would otherwise be useful, because even in high-variance regions, the RAW distributions tell us \emph{something} about the true scene color. This is especially true where the RAW distributions are strongly oriented (Fig.~\ref{fig:teaser} and bottom-left of Fig.~\ref{fig:ResponseFunc}): even though they have high total variance, most of their uncertainty is contained in one or two directions within RAW color space.

In this paper, we argue that vision systems can benefit substantially by incorporating a model of radiometric uncertainty when analyzing tone-mapped, JPEG-color images. We introduce a  probabilistic approach for visual inference, where (a) the calibrated estimate of a camera's forward tone-map is used to derive a probability distribution, for each tone-mapped JPEG color, over the RAW linear scene colors that could have induced it; and (b) the uncertainty embedded in these distributions is propagated to subsequent visual analyses. Using a variety of cameras and new formulations of a representative set of classic inference problems (multi-image fusion, photometric stereo, and deblurring), we demonstrate that modeling radiometric uncertainty is important for achieving optimal performance in computer vision.

The paper is organized as follows. After related work in Sec.~\ref{sec:related_work}, Sec.~\ref{sec:crm} reviews parametric forms for modeling the global tone-maps of consumer digital cameras and describes an algorithm for fitting model parameters to offline training data. In Sec.~\ref{sec:pinv}, we demonstrate how any forward tone-map model can be used to derive per-pixel inverse color distributions, that is, distributions for linear RAW colors conditioned on the JPEG color reported at each pixel. Section~\ref{sec:apps} shows how the uncertainty in these inverse distributions can be propagated to subsequent visual processes, by introducing new formulations of a representative set of classical inference tasks: image fusion (\eg, \cite{mann1995bud}); three-dimensional shape via Lambertian photometric stereo (\eg, \cite{photometric}); and removing camera shake via image deblurring (\eg, \cite{krishnan}).

\section{Related Work}
\label{sec:related_work}

The problem of radiometric calibration, where the goal is inverting non-linear distortions of scene radiance that occur during image capture and rendering, has received considerable attention in computer vision. Until recently, this calibration has been formulated only for grayscale images, or for color images on a per-channel-basis by assuming that the ``radiometric response function'' in each channel acts independently~\cite{mann1995bud, mitsunaga1999rsc, debevec97hdr, emor}. While early variants of this approach parametrized these response functions simply as an exponentiation (or ``gamma correction'') with the exponent as a single model parameter, later work sought to improve modeling accuracy by considering more general polynomial forms~\cite{emor}. Since these models have a relatively small number of parameters, they have featured in several algorithms for ``self-calibration''---parameter estimation from images captured in the wild, without calibration targets---through analysis of edge profiles~\cite{lin2004rcs,TaiCKLYYMB:2013}, image statistics~\cite{farid2002blind,kuthirummal2008plp}, or exposure stacks of images~\cite{mann1995bud,mitsunaga1999rsc,debevec97hdr,grossberg2003determining,reinhard2006high,shi2010self}.

However, per-channel models cannot accurately model the color processing pipelines of most consumer cameras, where the linear sensor measurements span a much wider gamut than the target output format. To be able to generate images that ``look good'' on limited-gamut displays, these cameras compress out-of-gamut and high-luminance colors in ways that are as pleasing as possible, for example by preserving hue. This means that two scene colors with the same raw sensor value in their red channels can have very different red values in their mapped JPEG output if one RAW color is significantly more saturated than the other.

Chakrabarti et al.~\cite{bmvc09} investigated the accuracy of more general, cross-channel parametric forms for global tone-mapping in a number of consumer cameras, including multi-variate polynomials and combinations of cross-channel linear transforms with per-channel polynomials. While they found reasonable fits for most cameras, the residual errors remained relatively high even though the calibration and evaluation were both limited to images of a single relatively narrow-gamut chart.  Kim et al.~\cite{kim2012new} improved on this by explicitly reasoning about the mapping of out-of-gamut colors. Their model consists of a cascade of: a linear transform, a per-channel polynomial, and a cross-channel correction for out-of-gamut colors using radial basis functions. The forward tone-map model we use in this paper (Sec.~\ref{sec:crm}) is strongly motivated by this work, although we find a need to augment the calibration training data so that it better covers the full space of measurable RAW values.

All of these approaches are focussed on modeling the distortion introduced by global tone-mapping. They do not, however, consider the associated loss of information, nor the structured uncertainty that exists when the distortion is undone as a pre-process for radiometric reasoning by vision systems. Indeed, while the benefit of undoing radiometric distortion has been discussed in the context of various vision applications (\eg, deblurring~\cite{ChenLYY:2012,TaiCKLYYMB:2013}, high-dynamic range imaging~\cite{pal2004probability}, video segmentation~\cite{GrundmannMKEK:2013}), previous methods have relied exclusively on deterministic inverse tone-maps that ignore the structured uncertainty evident in Figures~\ref{fig:teaser} and \ref{fig:ResponseFunc}. The main goal of this of this paper is to demonstrate that the benefits of undoing radiometric distortion can be made significantly greater by explicitly modeling the uncertainty inherent to inverse tone-mapping, and by propagating this uncertainty to subsequent visual inference algorithms.
 
 A earlier version of this work~\cite{xiong2012pixels} presented a direct method to estimate inverse RAW distributions from calibration data. In contrast, we introduce a two-step approach, where (a) calibrations images are used to fit the forward deterministic tone-map for a given camera, and (b) the model is inverted probabilistically. We find that this leads to better calibration and better inverse distributions with less calibration data.
 
Finally, we note that our proposed framework applies to stationary, global tone-mapping processes, meaning those that operate on each pixel independent of its neighboring pixels, and are unchanging from scene to scene. This is applicable to many existing consumer cameras locked into fixed imaging modes (``portrait'', ``landscape'' \etc), but not to local tone-mapping operators that are commonly used for HDR tone-mapping.

\section{Camera Rendering Model}
\label{sec:crm}

Before introducing our radiometric uncertainty model in Sec.~\ref{sec:pinv}-\ref{sec:apps}, we review and refine here a model for the forward tone-maps of consumer cameras, along with offline calibration procedures. We use a similar approach to Kim \etal~\cite{kim2012new}, and employ a two-step model to account for a camera's processing pipeline---a linear transform and per-channel polynomial, followed by a corrective mapping step for out-of-gamut and saturated colors. The end result is a deterministic forward map $\mathbb{J}: x\rightarrow y$ from 
RAW tricolor sensor measurements at a pixel $x \in [0,1]^3$ to corresponding rendered JPEG color values $y \in \{[0,255] \cap \mathbb{Z}\}^3$. Readers familiar with \cite{kim2012new} may prefer to skip directly to Sec.~\ref{sec:pinv}, where we present how to invert the modeled tone-maps probabilistically.

\subsection{Model}
\label{sec:model}

\begin{figure}[!t]
  \centering
  \includegraphics[width=0.9\columnwidth]{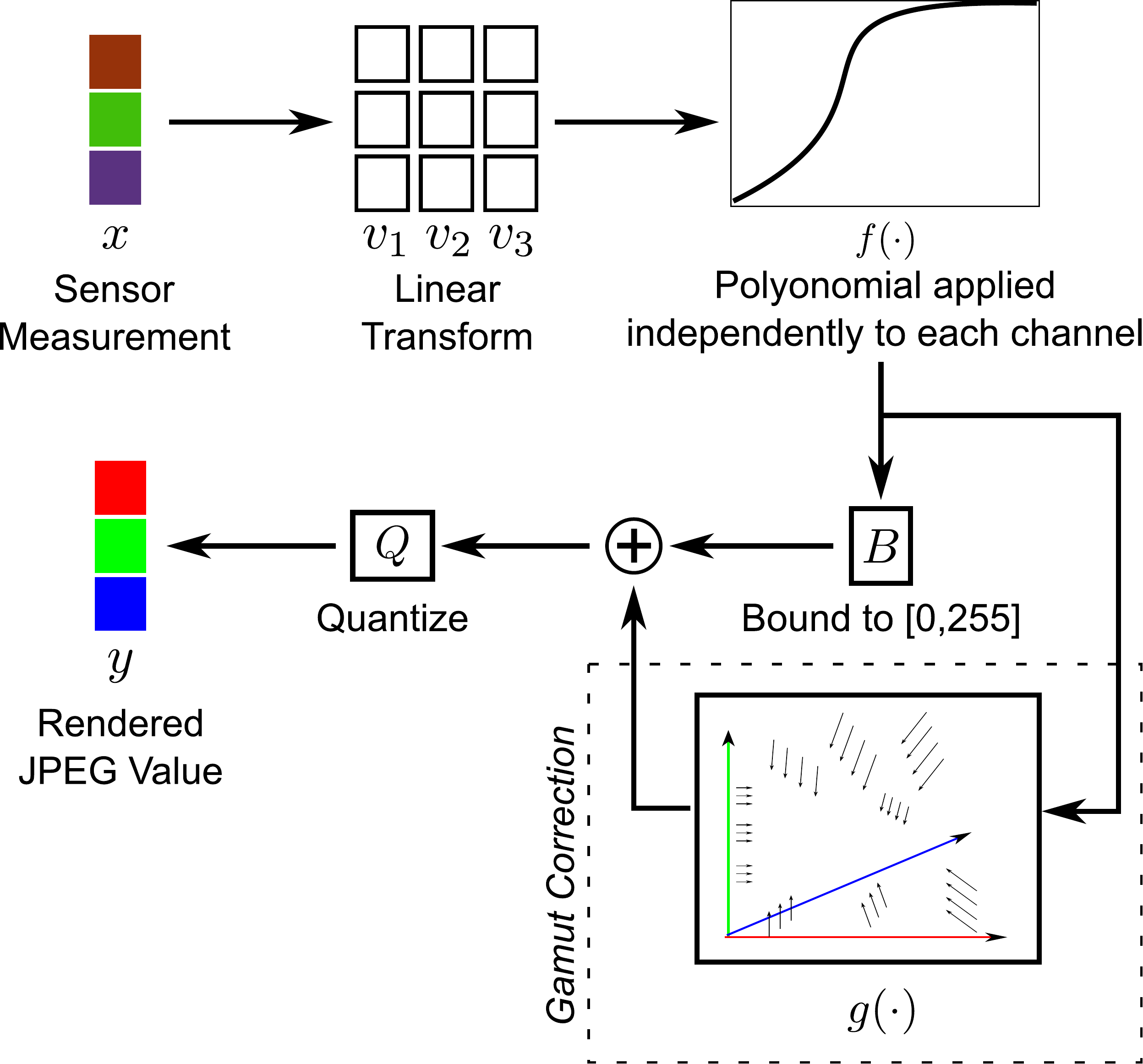}
  \caption{Rendering Model. We model a camera's processing pipeline using a two step-approach: (1) a $3\times 3$ linear transform and independent per-channel polynomial; followed by, (2) a correction to account for deviations in the rendering of saturated and out-of-gamut colors.}
  \label{fig:model}
\end{figure}

As shown in Fig.~\ref{fig:model}, we model the mapping $\mathbb{J}: x\rightarrow y$ as: 
\begin{eqnarray}
  \tilde{y} &=& \left[\begin{array}{c}\tilde{y}_1\\\tilde{y}_2\\\tilde{y}_3 \end{array}\right]
  = \left[\begin{array}{c}f(v_1^Tx)\\f(v_2^Tx)\\f(v_3^Tx) \end{array}\right],\label{eq:st}\\
  y &=& Q\left(B(\tilde{y}) + \left[\begin{array}{c} g_1(\tilde{y})\\ g_2(\tilde{y})\\g_3(\tilde{y})\end{array}\right] \right),\label{eq:gc}
\end{eqnarray}
where $v_1,v_2,v_3 \in \mathbb{R}^3$ define a linear color space transform, $B(\cdot)$ bounds its argument to the range $[0,255]$, and $Q(\cdot)$ quantizes its arguments to 8-bit integers. 

Equation (\ref{eq:st}) above corresponds to the commonly used per-channel polynomial model (\eg, \cite{emor,bmvc09}). Specifically, $f(\cdot)$ is assumed to be a polynomial of degree $d$:
\begin{equation}
  \label{eq:polydef}
  f(x) = \sum_{i=0}^d \alpha_i x^i,   
\end{equation}
where $\alpha_i$ are model parameters. We use seventh-order polynomials (\ie, $d=7$) in our implementation.

Motivated by the observations in \cite{kim2012new}, this polynomial model is augmented with an additive correction function $g(\cdot)$ in (\ref{eq:gc}) to account for deviations that result from camera processing to improve the visual appearance of rendered colors. We use support-vector regression (SVR) with a Gaussian radial basis function (RBF) kernel to model these deviations, \ie, each $g_c(\cdot), c \in \{1,2,3\}$ is of the form:
\begin{equation}
  \label{eq:rbf}
  g_c(\tilde{y}) = \sum_i \lambda_{c:i} \exp\left( - \gamma_c \|\tilde{y} - y_{c:i}\|^2 \right),
\end{equation}
where $\lambda_{c:i},y_{c:i}$ and $\gamma_c$ are also model parameters.

\subsection{Parameter Estimation}
\label{sec:calibration}

Next, we describe an algorithm to estimate the various parameters of this model from a set of calibration images. Using pairs of corresponding RAW-JPEG pixel values $\left\{(x_t,y_t)\right\}_{t=1}^T$ from the calibration set, we begin by estimating the parameters of the standard map in \eqref{eq:st} as:
\begin{equation}
  \label{eq:fit1}
  \{\hat{v}_c\},\{\hat{\alpha}_i\} = \argmin_{\{v_c\},\{\alpha_i\}}~ \sum_t  w_t \sum_c \|f(v_c^Tx_t)-y_{t:c} \|^2,
\end{equation}
where $\{w_t\}$ are scalar weights. Like \cite{bmvc09}, we also restrict $\{\alpha_i\}$ such that $f(\cdot)$ is monotonically increasing.

The weights $w_t$ are chosen with two objectives: (a) to promote a better fit for non-saturated colors, since we expect the corrective step in \eqref{eq:gc} to rectify rendering errors for saturated colors, and (b) to compensate for non-uniformity in the training set, \ie, more training samples in some regions over others. Accordingly, we set these weights as:
\begin{equation}
  \label{eq:wtdef}
  w_t = \mathcal{S}(y_t) \left[\sum_{t'=1}^T \exp\left(-\frac{|x_t-x_{t'}|^2}{2\sigma_t^2}\right)\right]^{-1},
\end{equation}
where $\mathcal{S}(y)$ is a scalar function that varies from $1$ to $0.01$ with increasing saturation in $y$, and the second term effectively re-samples the training set uniformly over the RAW space. We set $\sigma_t = T^{-1/3}$ to correspond to the expected separation between $T$ uniformly sampled points in the $[0,1]^3$ cube. 
\begin{table*}[!t]
  \centering
  \renewcommand{\arraystretch}{1.3}
  \caption{RMSE for Estimated Rendering Functions in Gray Levels for RAW-capable Cameras}
  \label{tab:fferrs}
  \begin{tabular}{c||c||c|c|c|c|c|c}\hline
    Camera Name & Uniform & 10 Exp. & 10 Exp. & 4 Exp. &\bf 8 Exp. & 4 Exp. & 8 Exp.\\
    & 8k Samples & 1 Illum. & 2 Illum. & 4 Illum. &\bf 4 Illum. & 6 Illum. & 6 Illum. \\\hline\hline
    
Panasonic DMC LX3&$1.50$&$6.64$&$5.51$&$2.98$&$2.37$&$2.56$&$2.41$\\ \hline
Canon EOS 40D&$1.77$&$9.54$&$9.00$&$3.98$&$2.66$&$3.05$&$2.25$\\ \hline
Canon PowerShot G9 &$1.90$&$10.44$&$3.77$&$3.24$&$2.51$&$2.96$&$2.60$\\ \hline
Canon PowerShot S90&$2.42$&$4.92$&$3.58$&$3.40$&$2.95$&$3.25$&$2.82$\\ \hline
Nikon D7000 &$1.65$&$10.29$&$3.69$&$23.49$&$2.21$&$2.52$&$2.03$\\ \hline
  \end{tabular}
\end{table*}

Once we have set the weights, we use an approach similar to the one in \cite{bmvc09} to minimize the cost in \eqref{eq:fit1}. We alternately optimize over only the linear or polynomial parameters, $\{v_c\}$ and $\{\alpha_i\}$ respectively, while keeping the other constant. For fixed $\{v_c\}$, the optimal $\{\alpha_i\}$ can be found by using a standard quadratic program solver, since the cost in \eqref{eq:fit1} is quadratic in $\{\alpha_i\}$, and the monotonicity restriction translates to linear inequality constraints. For fixed $\{\alpha_i\}$, we use gradient descent to find the optimal linear parameters $\{v_c\}$.

We begin the above alternating iterations by assuming $f(x) = x$ and setting $\{v_c\}$ directly using least-squares on training samples for which $y_t$ is small--- this is based on the assumption that $f(x)$ is nearly linear for small values of $x$. We then run the iterations till convergence, but since the cost in \eqref{eq:fit1} is not convex, there is no guarantee that the iterations above will yield the global minimum. Therefore, we restart the optimization multiple times with estimates of $\{v_c\}$ corresponding to random deviations around the current optimum.

Finally, we compute the parameters of the gamut mapping functions $\{g_c(\cdot)\}$ by using support-vector regression~\cite{libsvm} to fit $\tilde{y} \rightarrow \left[ y - C(\tilde{y}) \right]$, where the training samples $\{\tilde{y}_t\}$ are computed from $\{x_t\}$ using \eqref{eq:st} with the parameter values estimated above, and the kernel \emph{bandwidth} parameters $\{\gamma_c\}$ set using cross-validation.

\subsection{Datasets}
\label{sec:prac}

Our database consists of images captured using a number of popular consumer cameras (see Tables~\ref{tab:fferrs} and \ref{tab:fferrs2}), using an X-Rite 140-patch color checker chart as the calibration target as in \cite{bmvc09} and \cite{kim2012new}. However, although the chart contains a reasonably wide gamut of colors, these colors only span a part of the space of possible RAW values that can be measured by a camera sensor. 

To be able to reliably fit the behavior of each camera's tone-mapping function in the full space of measurable scene colors, and to accurately evaluate the quality of these fits, we captured images of the chart under \emph{sixteen} different illuminants (we used a standard Tungsten bulb paired with different commercially available gel-based color filters) to obtain a significantly wider gamut of colors. Moreover, for each illuminant, we captured images with different exposure values that range from one where almost all patches are under-exposed to one where all are over-exposed. We expect this collection of images to represent an exhaustive set that includes the full gamut of irradiances likely to be present in a scene.

Most of the cameras in our dataset allow access to the RAW sensor measurements, and therefore directly give us a set of RAW-JPEG pairs for training and evaluation. For JPEG-only cameras, we captured a corresponding set of images using a RAW-capable camera. To use the RAW values from the second camera as a valid proxy, we had to account for the fact that the exposure steps in the two cameras were differently scaled (but available from the image metadata), and for the possibility that the RAW proxy values in some cases may be clipped while those recorded by the JPEG camera's sensors were not. Therefore, the exposure stack for each patch under each illuminant from the RAW camera was used to estimate the underlying scene color at a canonical exposure value, and these were then mapped to the exposure values from the JPEG camera without clipping. 

For a variety of reasons, we expect the quality of fit to be substantially lower when using a RAW proxy. Our model does not account for the fact that there may be different degrees of vignetting in the two cameras, and it implicitly assumes that spectral sensitivity functions in the RAW and JPEG cameras are linearly related (\ie, that there is a bijective mapping between linear color sensor measurements in one camera and those in the other), which may not be these case~\cite{holm2006capture,jiang2013space}. Moreover, despite the fact that  the white balance setting in each camera is kept constant---we usually use ``daylight'' or ``tungsten''---we observe that some cameras exhibit variation in the white balance multipliers they apply for different scenes (different illuminants and exposures). For RAW-capable cameras, these multipliers are in the metadata and can be accounted for when constructing the calibration set. However, these values are not usually available for JPEG-only cameras, and thus introduce more noise in the calibration set.

\subsection{Evaluation}

For each camera, we estimated the parameters of our rendering model using different subsets of the collected RAW-JPEG pairs, and measured the quality of this calibration in terms of root mean-squared error (RMSE) values (between the predicted and true JPEG values, in terms of gray levels for an 8-bit image) on the entire dataset. These RMSE values for the RAW-capable camera are reported in Table.~\ref{tab:fferrs}. 

The first of these subsets is simply constructed with 8000 random RAW-JPEG pairs sampled uniformly across all pairs, and as expected, this yields the best results. Since capturing such a large dataset to calibrate any given camera may be practically burdensome, we also consider subsets derived from a limited number of illuminants, and with a limited number of exposures per-illuminant. The exposures are equally spaced from the lowest to the highest, and the subset of illuminants are chosen so as to maximize the diversity of included chromaticities--- specifically, we order the illuminants such that for each $n$, the convex hull of the RAW R-G chromaticities of patches from the first $n$ illuminants has the largest possible area. This order is determined using one of the cameras (the Panasonic DMC LX3), and used to construct subsets for all cameras. 

We find that different cameras show different degrees of sensitivity to diversity in exposures and illuminants, but using four illuminants with eight exposures represents a reasonable acquisition burden while also providing enough diversity for reliable calibration in all cameras. On the other hand, images of the chart under only a single illuminant, even with a large number of exposures, do not provide a diverse enough sampling of the RAW sensor space to yield good estimates of the rendering function across the entire dataset. 

Table~\ref{tab:fferrs2} shows RMSE values obtained from  calibrating JPEG-only cameras, and as expected, these values are substantially (approx.~3 to 4 times) higher than those for RAW-capable cameras.  Note that for this case, we only show results for the uniformly sampled training set, since we find parameter estimation to be unstable when using more limited subsets. This implies that calibrating JPEG-only cameras with a RAW proxy is likely to require the acquisition of larger sets of images, and perhaps more sophisticated fitting algorithms that explicitly infer and account for vignetting effects, scene-dependent variations in white balance, \etc.

\begin{figure}[!t]
  \centering
  \includegraphics[width=\columnwidth]{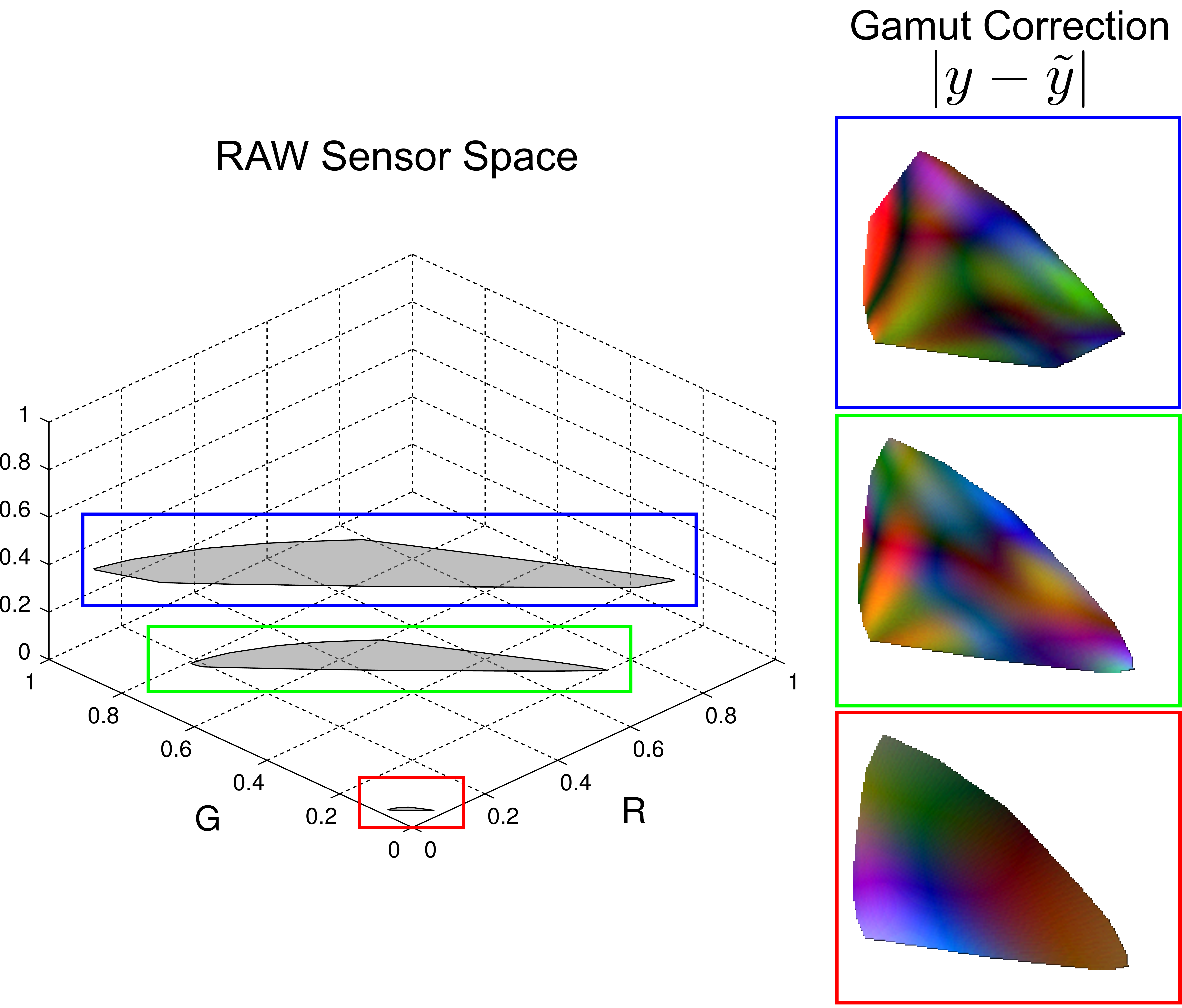}
  \caption{Estimated Gamut Correction Function for Canon PowerShot S90. For different slices of the RAW cube, we show the magnitude of the shift in each rendered JPEG channel (scaled by a factor of 8) due to gamut correction.
}
  \label{fig:ffview}
\end{figure}

Fig.~\ref{fig:ffview} illustrates the deviations due to the gamut correction step in our model, using the estimated rendering function for one of the calibrated cameras. We see that while this function is relatively smooth, its variations clearly can not be decomposed as per-channel functions. This confirms the observations in \cite{kim2012new} on the necessity of including a cross-channel correction function.

\begin{table}[!t]
  \centering
  \renewcommand{\arraystretch}{1.3}
  \caption{RMSE in Gray Levels for JPEG-only cameras}
  \label{tab:fferrs2}
  \begin{tabular}{c|c||c}\hline
    Camera Name & Raw Proxy & RMSE\\
    &&(w/ 8k Samples)\\\hline\hline
    Fujifilm J10 & Panasonic DMC LX3 & 10.24\\\hline
    Panasonic DMC LZ8 & Canon PowerShot G9 & 9.80\\\hline
    Samsung Galaxy S3 & Nikon D7000 & 11.47\\\hline
  \end{tabular}
 
\end{table}

\section{Probabilistic Inverse}
\label{sec:pinv}
\begin{table*}[!t]
  \centering
  \caption{Mean Empirical log-Likelihoods under Inverse Models for RAW-capable Cameras}
  \label{tab:elh}
  \renewcommand{\arraystretch}{1.3}
  \begin{tabular}{c||c||c|c|c|c}\hline
    \parbox[t]{8em}{~\vspace{-0.5em}\\Camera Name}\vspace{-0.5em} &Deterministic Inverse &Prob. Inverse&Prob. Inverse&Prob. Inverse&Prob. Inverse\\
    & Uniform 8k samples & Uniform 8k samples & 10 Exp., 2 Illum. & 4 Exp., 4 Illum. &\bf 8 Exp., 4 Illum.\\\hline\hline
    Panasonic DMC LX3 & 3.50 & 12.44 & 6.19 & 11.87 & 12.17 \\\hline
    Canon EOS 40D & 3.45 & 13.06 & -0.18 & 11.87 & 12.22 \\\hline
    Canon PowerShot G9 & 2.01 & 8.33 & 7.12 & 7.80 & 8.16\\\hline
    Canon PowerShot S90 & 3.83 & 11.34 & 10.47 & 10.96 & 10.91\\\hline
    Nikon D7000 & 1.59 & 8.52 & 6.20 & 3.45 & 8.28 \\\hline
  \end{tabular}
\end{table*}

The previous section dealt with computing an accurate estimate of tone-mapping function applied by a camera. However, the main motivation for calibrating a camera is to be able to invert this tone-map and use available JPEG values back to derive radiometrically meaningful RAW measurements that are useful for computer vision applications. But it is easy to see that this inverse is not uniquely defined, since multiple sensor measurements can be mapped to the same JPEG output as a result of the quantization that follows the compressive map in \eqref{eq:gc}, with higher intensities and saturated colors experiencing greater compression, and therefore more uncertainty in their recovery from reported JPEG values.

Therefore, instead of using a deterministic inverse function, we define the inverse probabilistically as a \emph{distribution} $p(x|y)$ of possible RAW measurements $x$ that could have been tone-mapped to a given JPEG output $y$. While formulating this distribution, we also account for errors in the estimate $\mathbb{J}$ of the rendering function, treating them as Gaussian noise with variance $\sigma_f^2$, where $\sigma_f$ is set to twice the \emph{in-training} RMSE. Specifically, we define $p(x|y)$ as:
\begin{equation}
  \label{eq:pxdef}
  p(x | y) = \frac{1}{Z}\,p(x)\exp\left(-\frac{\left\|y - \mathbb{J}(x) \right\|^2}{2\sigma_f^2}\right),
\end{equation}
where $Z$ is the normalization factor
\begin{equation}
  \label{eq:zdef}
  Z = \int p(x')\exp\left(-\frac{\left\|y - \mathbb{J}(x') \right\|^2}{2\sigma_f^2}\right)\,dx',
\end{equation}
and $p(x)$ is a \emph{prior} on sensor-measurements. This prior can range from per-pixel distributions that assert, for example, that broadband reflectances are more likely than saturated colors; to higher-order scene-level models that reason about the number of distinct chromaticities and materials in a scene--- we expect that the choice of $p(x)$ will be different for different applications and environments. In this paper, we simply choose a \emph{uniform} prior over all possible sensor measurements whose chromaticities lie in the convex hull of the training data.

Note that these distributions are computed assuming that the white balance multipliers are known (and incorporated in $\mathbb{J}$). For some cameras, even with a fixed white-balance setting, the actual white-balance multipliers might vary from scene to scene. In these cases, the variable $x$ in the distribution above will be a linear transform\footnote{Note that white-balance correction is typically a linear diagonal transform in the camera's sensor space. For cameras that are not RAW-capable and have been calibrated with a RAW proxy, this will be a general linear transform in the proxy's sensor space.}--- which is fixed for all pixels in a given image--- away from a scene-independent RAW measurement. This may be sufficient for applications that only reason about colors in a single image, or in multiple images of the same scene where the white balance multipliers can reasonably be expected to remain fixed, but other applications will need to address this ambiguity when using these inverse distributions.

While the expression in \eqref{eq:pxdef} is the exact form of the inverse distribution---corresponding to a uniform distribution over all RAW values $x$ predicted by the camera model to map to a given JPEG value $y$, with added slack for calibration error---it has no convenient closed form. Practitioners  will therefore need to compute them explicitly over a grid of possible values of $x$ for each JPEG value $y$, or approximate them with a convenient parametric form for use in vision applications. We employ multi-variate Gaussian distributions to approximate the exact form in \eqref{eq:pxdef}, as an example to demonstrate the benefits of using a probabilistic inverse in the remainder of this paper, but this is only one possible choice and the optimal representation for these distributions will likely depend on the target application and platform.

Formally, we approximate $p(x|y)$ as
\begin{eqnarray}
  \label{eq:musdef}
  \tilde{p}(x|y) &=& \mathcal{N}(x|\mu(y),\Sigma(y)),\notag\\
\mu(y) &=& \int x p(x|y)\, dx,\notag\\
  \Sigma(y) &=& \int \left(x - \mu(y)\right) \left(x - \mu(y)\right)^T p(x | y)\, dx.
\end{eqnarray}
Note that here $\mu$, in addition to being the mean of the approximate Gaussian distribution, is \emph{also} the single best estimate of $x$ given $y$ (in the minimum least-squares error sense) from the exact distribution in \eqref{eq:pxdef}. And since \eqref{eq:pxdef} is derived using a camera model similar to that of \cite{kim2012new}, $\mu$ can be interpreted as the deterministic RAW estimate that would be yielded by the algorithm in \cite{kim2012new}.

The integrations in \eqref{eq:musdef} are performed numerically, and by storing pre-computed values of $\mathbb{J}$ on a densely-sampled grid to speed up distance computations. A MATLAB implementation is available on our project page~\cite{projectpage}, which takes roughly $15ms$ to compute the mean and co-variance above for a single JPEG observation on a modern machine.

Tables~\ref{tab:elh} and \ref{tab:elh2} report the mean empirical log-likelihoods, \ie, the mean value of $\log \tilde{p}(x|y)$ across all RAW-JPEG pairs $(x,y)$ in the validation set, for our set of calibrated cameras. For the RAW-capable cameras, we report these numbers for inverse distributions computed using estimates of the rendering function $\mathbb{J}$ from different calibration sets as in Table~\ref{tab:fferrs}. As expected, better estimates of $\mathbb{J}$ usually lead to better estimates of $\tilde{p}$ with higher log-likelihoods, and  we find that our choice of calibrating using 8 exposures and 4 illuminants for RAW cameras yields scores that are close to those achieved by random samples across the entire validation set. 

Moreover, to demonstrate the benefits of using a probabilistic inverse, we also report log-likelihood scores from a deterministic inverse that outputs single prediction ($\mu$ from \eqref{eq:musdef}) for the RAW value for a given JPEG. Note that strictly speaking, the log-likelihood in this case would be $-\infty$ unless $\mu$ is exactly equal to $x$. The scores reported in Tables~\ref{tab:elh} and \ref{tab:elh2} are therefore computed by using a Gaussian distribution with variance equal to the mean prediction error (which is the choice that yields the maximum mean log-likelihood). We find that these scores are much lower than those from the full model, demonstrating the benefits of a probabilistic approach.

\begin{figure*}[!t]
  \centering
  \rotatebox{90}{~~~~~~~~~~~~~~~~~\footnotesize Canon PowerShot S90}  
 \includegraphics[width=0.98\columnwidth,trim=9em 6em 7em 6em,clip]{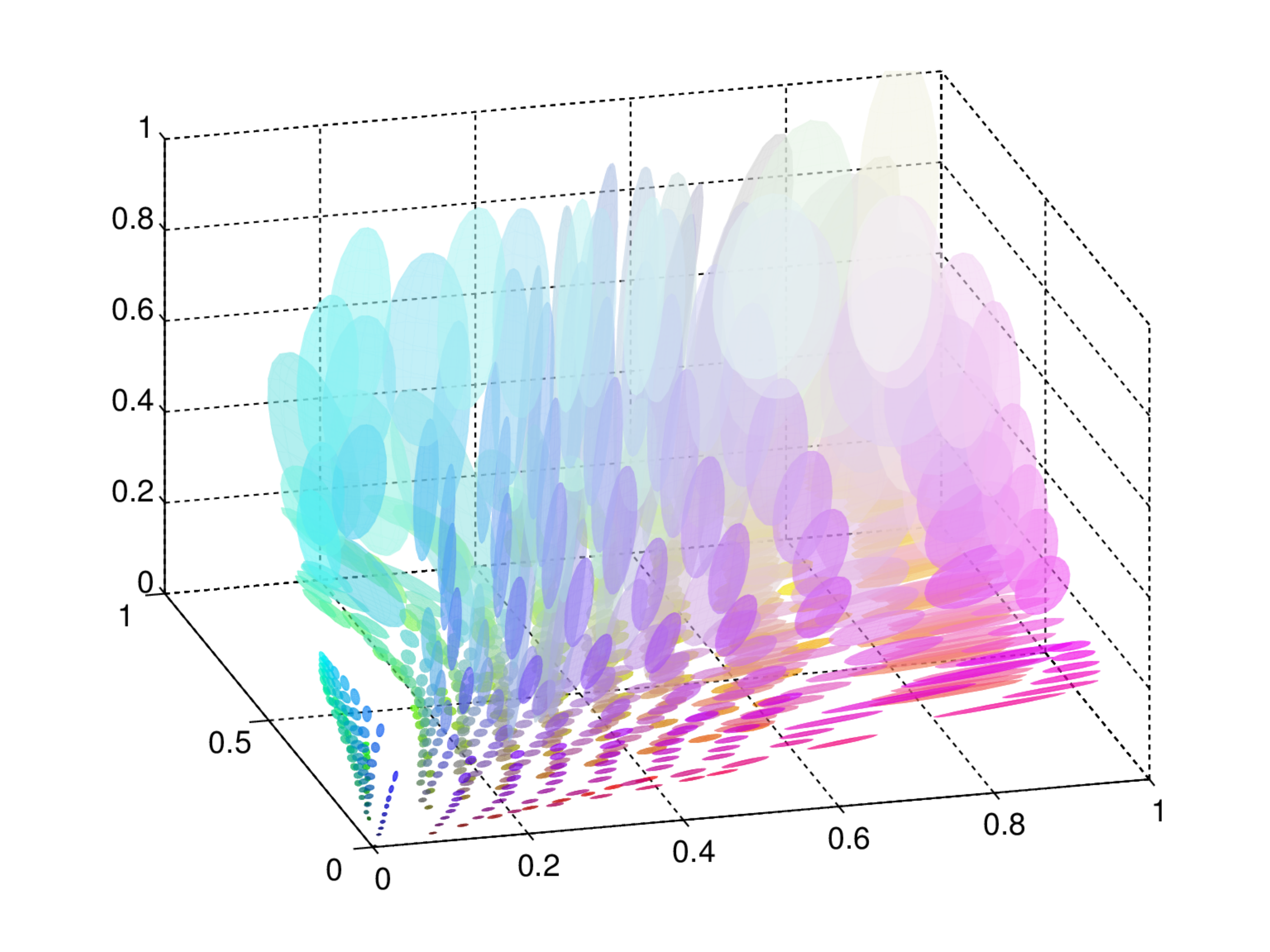}
 \rotatebox{90}{~~~~~~~~~~~~~~~~~~~~~~\footnotesize Canon EOS 40D} \includegraphics[width=0.98\columnwidth,trim=9em 6em 7em 6em,clip]{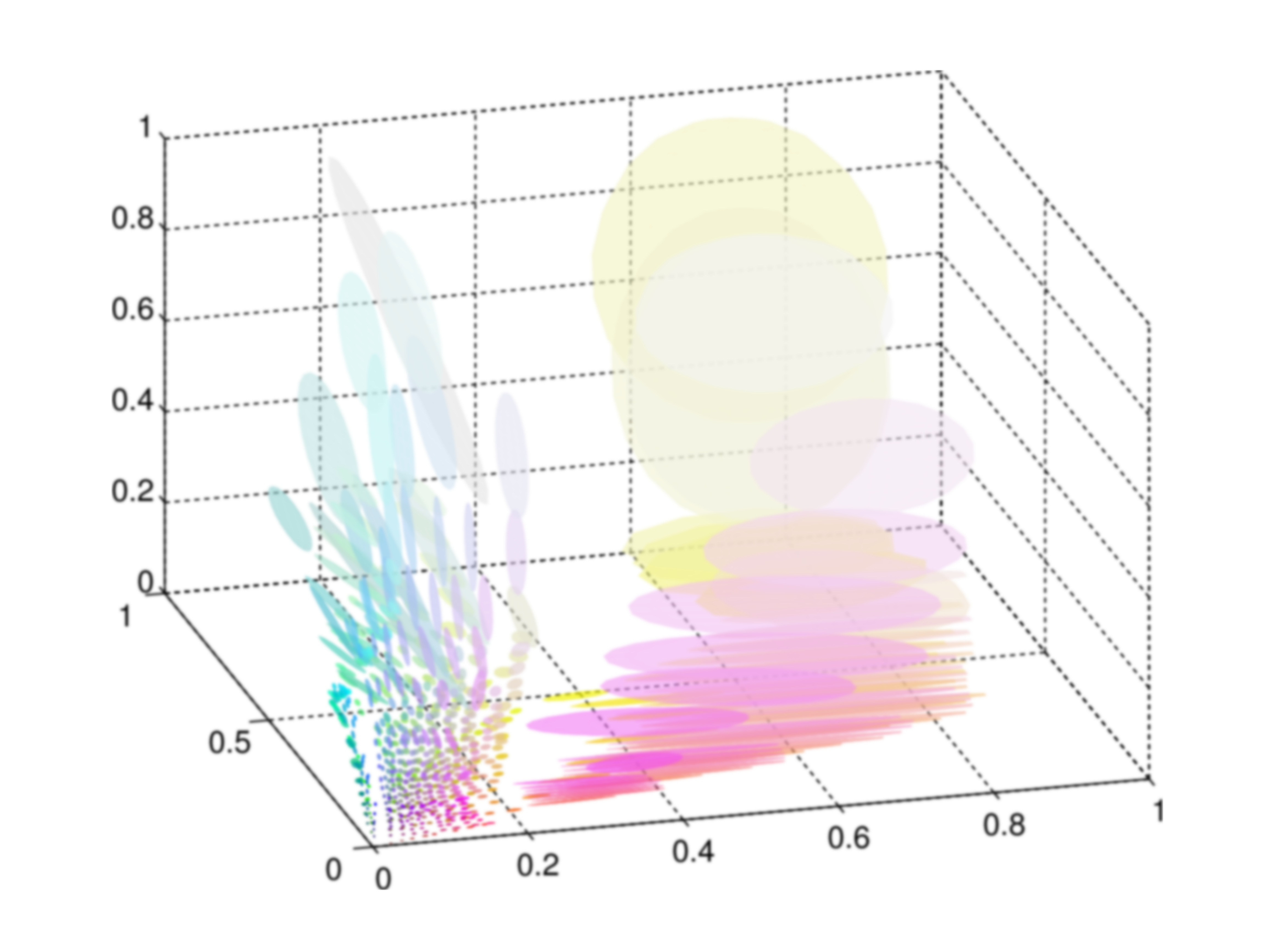}\\
 \rotatebox{90}{~~~~~~~~~~~~~~~~~\footnotesize Canon PowerShot G9} \includegraphics[width=0.98\columnwidth,trim=9em 6em 7em 6em,clip]{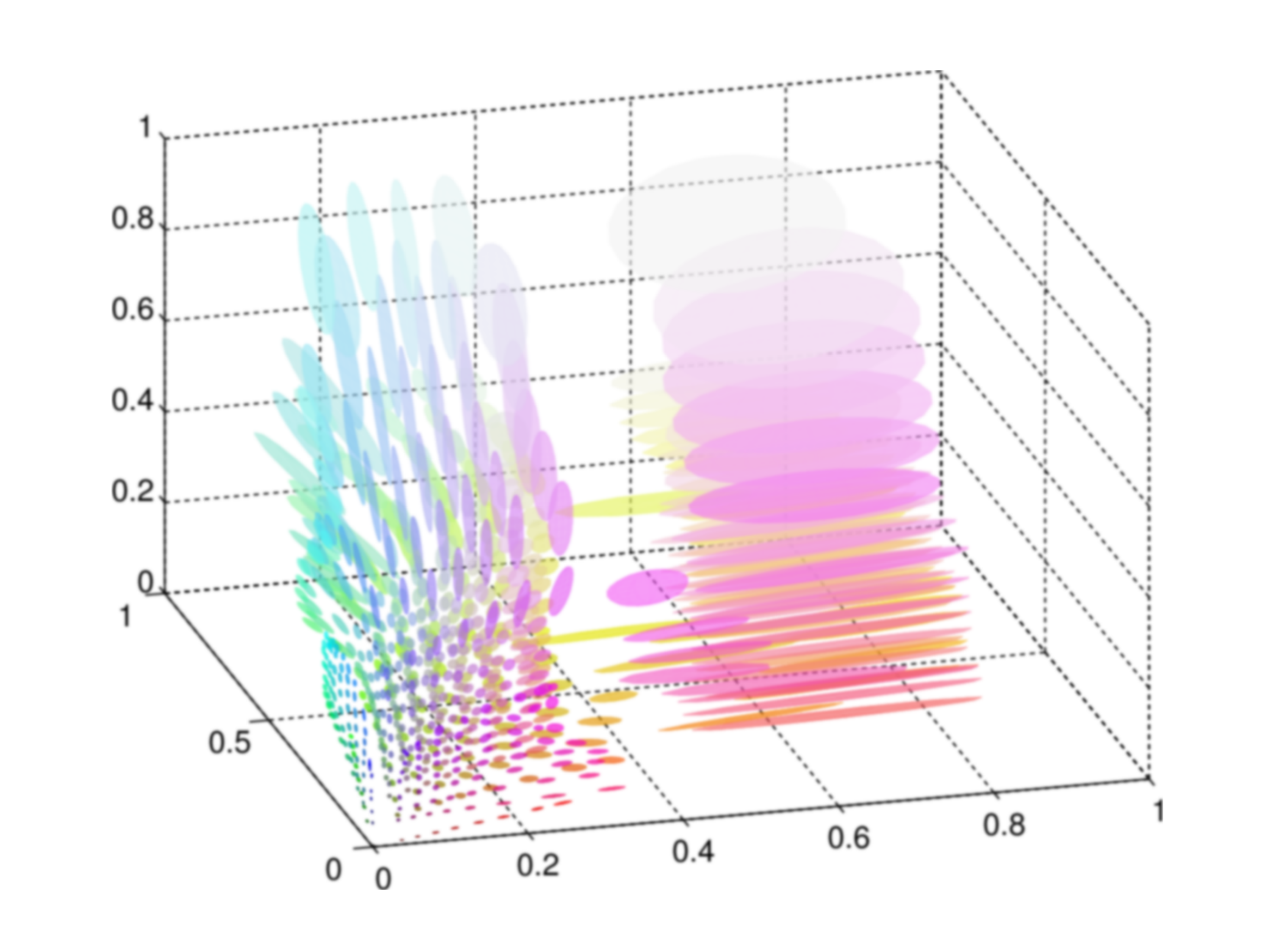}
 \rotatebox{90}{~~~~~~~~~~~~~~~~~~~~~~\footnotesize Nikon D7000} \includegraphics[width=0.98\columnwidth,trim=9em 6em 7em 6em,clip]{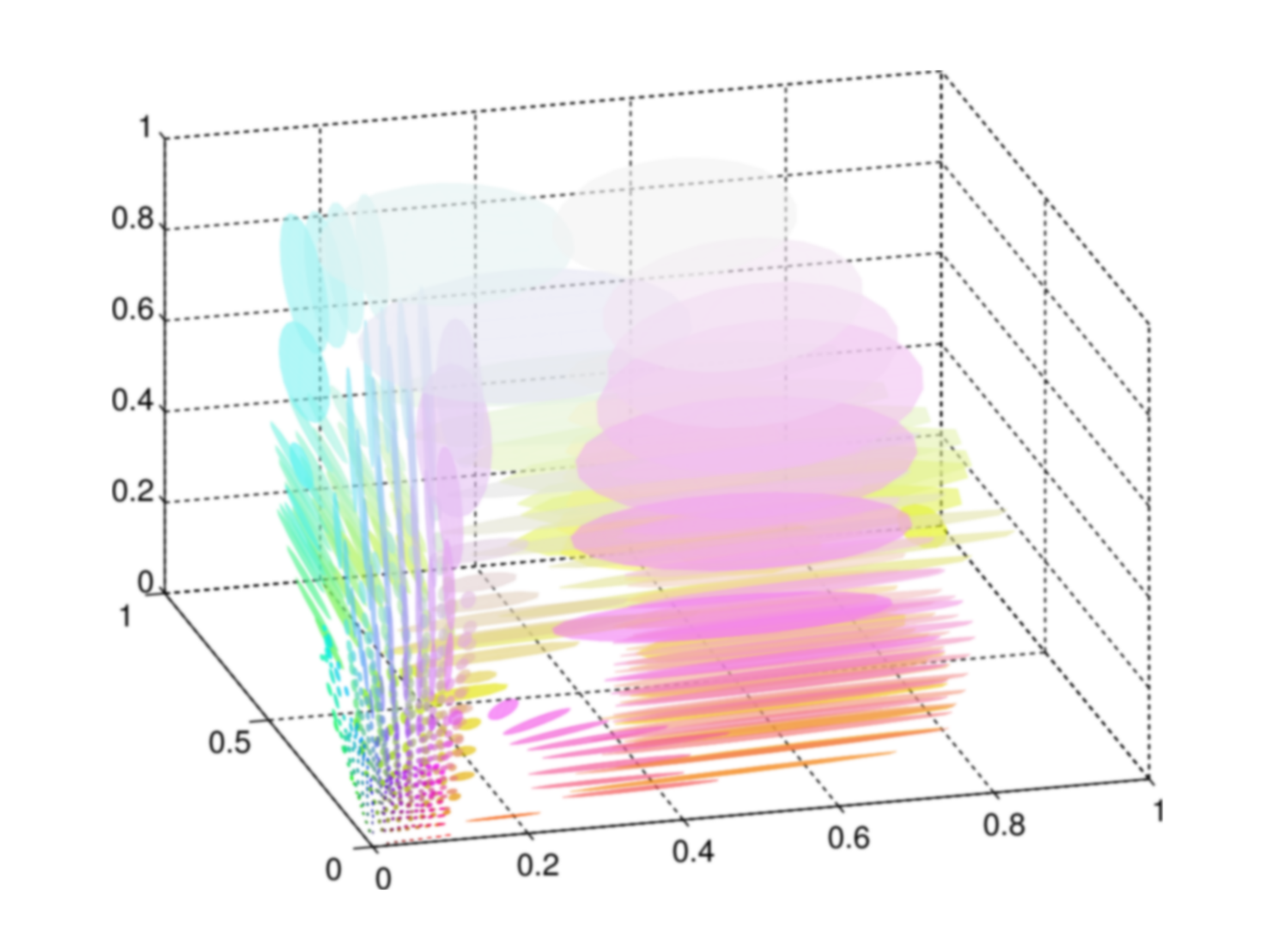}\\

\caption{Probabilistic Inverse. For different cameras, we show ellipsoids in RAW space that denote  the mean and covariance of $p(x|y)$ for different JPEG values $y$---indicated by the color of the ellipsoid. These values $y$ are uniformly sampled in JPEG space, and we note that the corresponding distributions can vary significantly across cameras.}
  \label{fig:piview}
\end{figure*}

Finally, we show visualizations of the inverse distributions for four of the remaining RAW-capable cameras in our database. These plots represent the distributions $\tilde{p}(x|y)$ using ellipsoids to represent mean and covariance, and can be interpreted as RAW values that are likely to be mapped to the same JPEG color by the camera. We see that these distributions are qualitatively different for different cameras, since different manufacturers typically employ their own strategies for compressing wide gamut sensor measurements to narrow gamut images that are visually pleasing. Moreover, the sizes and orientations of the covariance matrices can also vary significantly for different JPEG values $y$ obtained from the same camera.

\begin{table}[!t]
  \centering
  \caption{Mean Empirical log-Likelihoods for JPEG-only Cameras}
  \label{tab:elh2}
  \renewcommand{\arraystretch}{1.3}
  \begin{tabular}{c||c|c}\hline
    Camera Name & Deterministic Inverse & Prob. Inverse\\\hline\hline
    Fujifilm J10 & 1.97 & 8.69\\\hline
    Panasonic DMC LZ8 & 1.60 & 11.83\\\hline
    Samsung Galaxy S3 & 2.23 & 7.51\\\hline
  \end{tabular}\vspace{-0.5em}%
\end{table}

\section{Visual Inference with Uncertainty}
\label{sec:apps}

The probabilistic derendering model \eqref{eq:musdef} provides an opportunity for vision systems to exploit the structured uncertainty that is unavoidable when inverting global tone-mapping processes. 
To demonstrate how vision systems can benefit from modeling this uncertainty, we introduce inference algorithms that incorporate it for a broad, representative set of visual tasks: image fusion, photometric stereo, and deblurring.

\subsection{Image Fusion}
\label{sec:hdr}

We begin with the task of combining multiple color observations of the same scene to infer accurate estimates of scene color. This task is essential to high dynamic-range (HDR) imaging from exposure-stacks of JPEG images in the spirit of Debevec and Malik~\cite{debevec97hdr}; and variations of it appear when stitching images together for harmonized, wide-view panoramas or other composites, and when inferring object color (intrinsic images and color constancy) or surface BRDFs from Internet images.

Formally, we consider the problem of estimating the linear color $x$ of a scene point from multiple JPEG observations $\{y_i\}$ captured at known exposures $\{\alpha_i\}$. Each observation $y_i$ is assumed to be the rendered version of sensor value $\alpha_i x$, and we assume the camera has been pre-calibrated as described previously. The naive extension of RAW HDR reconstruction is to use a deterministic approach to derender each JPEG value $y_i$, and then compute scene color $x$ using least-squares. This strategy considers every derendered JPEG value to be equally reliable and is implicit, for example, in traditional HDR algorithms based on self-calibration from non-linear images~\cite{mann1995bud,mitsunaga1999rsc,debevec97hdr,grossberg2003determining,reinhard2006high,shi2010self}. When the imaging system is pre-calibrated, the deterministic approach corresponds to ignoring variance information and computing a simple, exposure-corrected linear combination of the derendered means:
\begin{equation}
  \label{eq:hdru}
  x = \arg \min_x \sum_i \|\mu_i - \alpha_i x\|^2 = \frac{\sum_i \alpha_i \mu_i}{\sum_i \alpha_i^2},
\end{equation}
where $\mu_i = \mu(y_i)$.  

\begin{figure*}[!t]
  \centering
  \includegraphics[width=\textwidth]{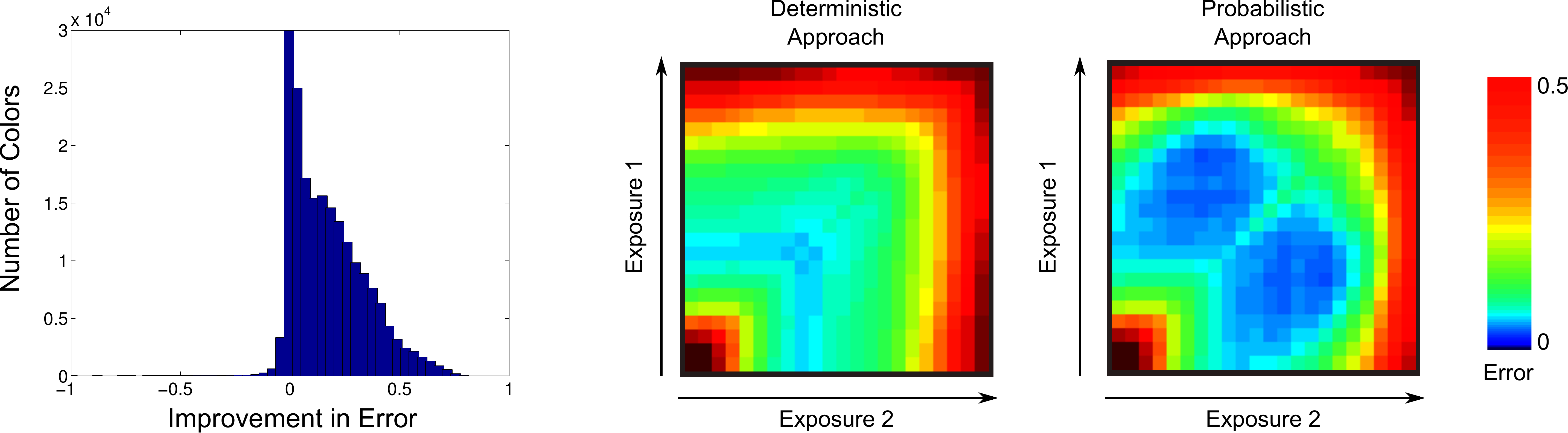}
  \caption{HDR Results on the Panasonic DMC LX3. (Left) Histogram of improvement in errors over the deterministic baseline for all scene colors using every possible exposure pair. (Right) Mean errors across all colors for each approach when using different exposure pairs.}
  \label{fig:hdr}
\end{figure*}

In contrast to the deterministic approach, we propose using the probabilistic inverse from Sec.~\ref{sec:pinv} to weigh the contribution of each JPEG observation based on its reliability, thereby improving estimation. Estimation is also improved by the fact that inverse distributions from different exposures of the same scene color often carry complementary information, in the form of differently-oriented covariance matrices. Specifically, each observation provides us with a Gaussian distribution $p(x|y_i,\alpha_i) = \mathcal{N}(x|\mu_i,\Sigma_i)$,
\begin{equation}
  \label{eq:hdrp}
  \Sigma_i = \frac{\Sigma(y_i) + \sigma_z^2 I_{3\times 3} }{\alpha_i^2},
\end{equation}
where $\sigma_z^2$ corresponds to the expected variance of photo-sensor noise, which is assumed to be constant and small relative to most $\Sigma_i$. The most-likely estimate of $x$ from all observations is then given by
\begin{align}
  x &= \arg \max_{x} \prod_i \mathcal{N}(x|\mu_i,\Sigma_i)\notag\\& = \left(\sum_i \Sigma_i^{-1} \right)^{-1} \left(\sum_i \Sigma_i^{-1}\mu_i\right).
\end{align}

An important effect that we need to account for in this probabilistic approach is clipping in the photo-sensor. To handle this, we insert a check on the derendered distributions $\left(\mu(y_i),\Sigma(y_i)\right)$, and when the estimated mean in any channel is close to $1$, we update the corresponding elements of $\Sigma_i$ to reflect a very high variance for that channel. The same strategy is also adopted for the baseline deterministic approach~\eqref{eq:hdru}.

To experimentally compare reconstruction quality of the deterministic and probabilistic approaches, we use all RAW-JPEG color-pairs from the database of colors captured with the Panasonic DMC LX-3, corresponding to all color-pairs except those from the four training illuminants. We consider the color checker under a particular illuminant to be the target HDR scene, and we consider the differently-exposed JPEG images under that illuminant to be the input images of this scene. The task is to estimate for each target scene (each illuminant) the true linear patch color from only two differently-exposed JPEG images. The true linear patch color for each illuminant is computed using RAW data from all exposures, and performance is measured using relative RMSE:
\begin{equation}
  \label{eq:erdef}
  \mbox{Error}(x,x_{\mbox{\tiny true}}) = \frac{\|x - x_{\mbox{\tiny true}}\|}{\|x_{\mbox{\tiny true}}\|}.
\end{equation}

Figure~\ref{fig:hdr} shows a histogram of the reduction in RMSE values when using the probabilistic approach. This is the histogram of differences between evaluating \eqref{eq:erdef} with probabilistic and deterministic estimates $x$ across $1680$ distinct linear scene colors in the dataset and all possible un-ordered pairs of $22$ exposures\footnote{These correspond to the different exposure time stops available on the camera: $[5e{-4},$ $6.25e{-4},$ $1e{-3},$ $1.25e{-3},$ $2e{-3},$ $2.5e{-3},$ $3.13e{-3},$ $5e{-3},$ $6.25e{-3},$ $1e{-2},$ $1.26e{-2},$ $1.67e{-2},$ $2e{-2},$ $2.5e{-2},$ $3.33e{-2},$ $4e{-2},$ $5e{-2},$ $6.67e{-2},$ $1e{-1},$ $2e{-1},$ $4e{-1},$ $1]$ in relative time units.} as input, excluding the trivial pairs for which $\alpha_1=\alpha_2$ (a total of $388080$ test cases). In a vast majority of cases, incorporating derendering uncertainty leads to better performance. 

We also show in the right of the figure, for both the deterministic and probabilistic approaches, two-dimensional visualizations of the error for each exposure-pair. Each point in these visualizations corresponds to a pair of input exposure values $(\alpha_1, \alpha_2)$, and the pseudo-color depicts the mean RMSE across all $1680$ linear scene colors in the test dataset. (Diagonal entries correspond to estimates from a single exposure, and are thus identical for the probabilistic and deterministic approaches). We see that the probabilistic approach yields acceptable estimates with low errors for a larger set of exposure-pairs. Moreover, in many cases it leads to lower error than those from either exposure taken individually, demonstrating that the probabilistic modeling is not simply selecting the better exposure, but in fact combining complementary information from both observations.

\subsection{Photometric Stereo}

\label{sec:photometric_stereo}

Another important class of vision algorithms include those that deal with recovering scene depth and geometry. These algorithms are especially dependent on having access to radiometrically accurate information and have therefore been applied traditionally to RAW data, but the ability to reliably use tone-mapped JPEG images, say from the Internet, is useful for applications like weather recovery~\cite{shen2009photometric}, geometric camera calibration~\cite{lalonde2010sun}, and 3D reconstruction via photometric stereo~\cite{ackermann2012photometric}. As an example,  we consider the problem of recovering shape using photometric stereo from JPEG images.

Photometric stereo is a technique for estimating the surface normals of a Lambertian object by observing that object under different lighting conditions and a fixed viewpoint~\cite{photometric}. Formally, given images under $N$ different directional lighting conditions, with $ l_{i}\in\mathbb{R}^{3}$ being the direction and strength of the $i^{th}$ source, let $I_i\in\mathbb{R}$ denote the \emph{linear} intensity recorded in a single channel at a particular pixel under the $i^{th}$ light direction. If $\nu\in\mathbb{S}^{2}$ and $\rho\in\mathbb{R}^{+}$ are the normal direction and albedo of the surface patch at the back-projection of this pixel, then the Lambertian reflectance model provides the relation $\rho\langle l_{i},\nu\rangle=I_{i}$. The goal of photometric stereo is to infer the material $\rho$ and shape $\nu$ given the set $\{ l_{i},I_{i}\}$.

Defining a pseudo-normal $ b\triangleq\rho\nu$, the relation between the observed intensity and the scene parameters becomes
\begin{equation}
 l_{i}^{T} b=I_{i}.\label{eq:lr=00003DI}
\end{equation}
Given three or more $\{ l_{i},I_{i}\}$-pairs, the pseudo-normal $b$ is estimated simply using least-squares as:
\begin{equation}
  \label{eq:lsqps}
 b=( L^{T} L)^{-1} L^{T}I,
\end{equation}
where $L\in\mathbb{R}^{N\times 3}$ and $I\in\mathbb{R}^{N}$ are formed by stacking the light directions $ l_{i}^T$ and measurements $I_{i}$ respectively. The normal $\nu$ can then simply be recovered as $\nu = b/\|b\|$.

When dealing with a linear color image, Barsky and Petrou~\cite{barsky2001colour} suggest constructing the observations $I_i$ as a linear combination $I_{i}=c^T x_{i}$ of the different channels of the color vectors $x_i$. The coefficients $c\in \mathbb{R}^3$ are chosen to maximize the magnitude of the intensity vector $I$, and therefore the stability of the final normal estimate $\mu$, as
\begin{align}
c &= \arg\max_c~ \sum_i I_i^2 = \arg\max_c~ \sum_i \|c^T x_{i}\|^2,\notag\\
&= \arg \max_c~ c^T \left(\sum_i x_ix_i^T\right) c,~~\textrm{s.t.}\ \|c\|^2=1.
\end{align}
The optimal $c$ is then simply the eigenvector associated with the largest eigenvalue of the matrix $(\sum_i x_ix_i^T)$. Intuitively, this corresponds to the normalized color of the material at that pixel location.

In order to use photometric stereo to recover scene depth from JPEG images, we need to first obtain estimates of the linear scene color measurements $x_i$ from the available JPEG values $y_i$. Rather than apply the above algorithm as-is to deterministic inverse-mapped estimates of $x_i$, we propose a new algorithm that uses the distributions $p(x_i|y_i) = {\mathcal N}(x_i|\mu_i, \Sigma_i)$ derived in Sec.~\ref{sec:pinv}. 

First, we modify the approach in \cite{barsky2001colour} to estimate the coefficient vector $c$ by maximizing the signal-to-noise ratio (SNR), rather than simply the magnitude, of $I$:
\begin{align}
c &= \arg \max_c~\frac{\sum_i \mathbb{E}[I_i^2]}{\sum_i \mbox{Var}(I_i)}=\arg \max_c~\frac{\sum_i \textrm{E}[(c^T x_{i})^2]}{\sum_i \textrm{Var}[c^T x_{i}]}\notag\\
&= \arg \max_c~\frac{c^T\left(\sum_i \mu_i\mu_i^T + \Sigma_i\right) c}{c^T\left(\sum_i \Sigma_i \right)c}
~~~\textrm{s.t.}\ \|c\|^2=1.
\end{align}
It is easy to show that the optimal value of $c$ for this case is given by the eigenvector associated with the largest eigenvalue of the matrix $\left(\sum_i \Sigma_i\right)^{-1}\left(\sum_i \mu_i \mu_i^T\right)$. This choice of $c$ essentially minimizes the relative uncertainty in the set of observations $I_i=c^Tx_i$, which are now described by univariate Gaussian distributions:
\begin{equation}
 I_i\sim{\cal N}(m_i, \sigma_i^2)={\cal N}( c^T\mu_i,  c^T\Sigma_i c).
\end{equation}

From this it follows (\eg, \cite{elementOfStatisticalLearning}) that the maximum likelihood estimate of the pseudo-normal $b$ is obtained through weighted least-squares, with weights given by the reciprocal of the variance, \ie,
\begin{equation}
 b=( L^{T}W L)^{-1} L^{T}W m,
\end{equation}
where $m\in\mathbb{R}^N$ is constructed by stacking the means $m_i$, and $W=\textrm{diag}\{\sigma_{i}^{-2}\}_{i=1}^{N}$.

\begin{figure*}[!t]
\centering
\includegraphics[width=0.9\textwidth]{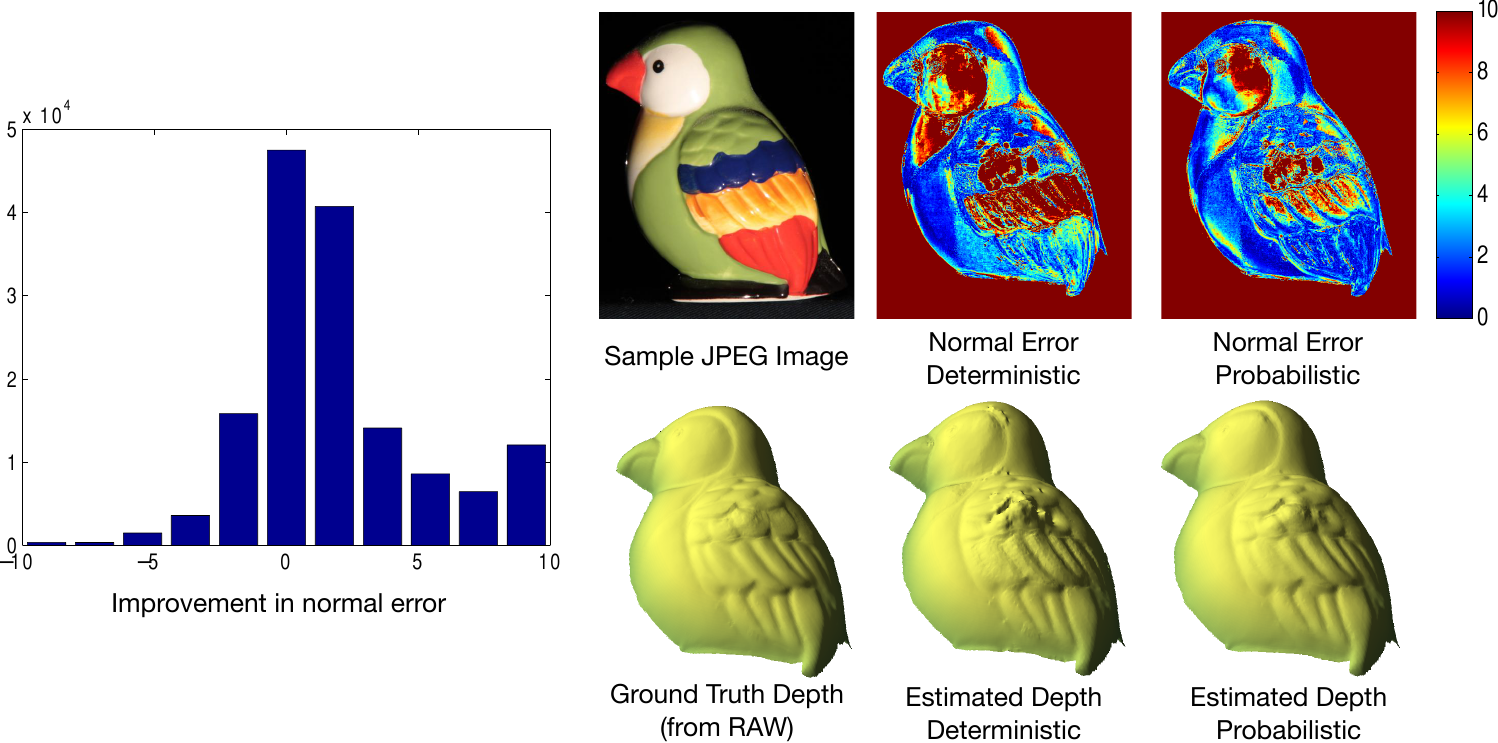}
\caption{\label{fig:photometricResult} Photometric Stereo Results using the Canon EOS 40D. (Left) Histogram of the improvement in angular error of normal estimate. (Right) One of the JPEG images used during estimation, and angular error (in degrees) for the normals estimated using the deterministic and probabilistic approaches, along with the corresponding depth maps.}
\end{figure*}

We evaluate our algorithm on JPEG images of a figurine captured using the Canon~EOS~40D  from a fixed viewpoint under directional lighting from ten different known directions. At each pixel, we discard the brightest and darkest measurements to avoid possible specular highlights and shadows, and use the rest to estimate the surface normal. The camera takes RAW images simultaneously, which are used to recover surface normals that we treat as ground truth.

Figure~\ref{fig:photometricResult} shows the angular error map for normal estimates using the proposed method, as well as the deterministic baseline. We also show the corresponding depth maps obtained from the normal estimates using \cite{frankotIntegration}. The proposed probabilistic approach produces smaller normal estimate errors and fewer reconstruction artifacts than the deterministic algorithm---quantitatively, the mean angular error is $4.34^{\circ}$ for the probabilistic approach, and $6.46^{\circ}$ for the deterministic baseline. We also ran the reconstruction algorithm on inverse estimates computed by simple gamma-correction on the JPEG values (a gamma parameter of 2.2 is assumed). These estimates had a much higher mean error $14.65^{\circ}$.

\subsection{Deconvolution}

Deblurring is a common image restoration application and has been an active area of research in computer vision~\cite{dblr0,dblr1,dblr2,dblr3,whyte2012non}. Traditional deblurring algorithms are designed to work on linear RAW images as input, but in most practical settings, only camera rendered JPEG images are available. The standard practice in such cases has been to apply an inverse tone-map assuming a simple gamma correction of $2.2$, but as has been recently demonstrated~\cite{kimdb}, this approach is inadequate and will often yield poor quality images with visible artifacts due to the fact that deblurring algorithms rely heavily on linearity of the input image values. 

While Kim \etal~\cite{kimdb} show that more accurate inverse maps can improve deblurring performance, their maps are still \emph{deterministic}. In this section, we explore the benefits of using a probabilistic inverse, and introduce a modified deblurring algorithm that accounts for varying degrees of uncertainty in estimates of RAW values from pixel to pixel.

Formally, given a blurry JPEG image $y(n)$, we assume that the corresponding blurry RAW image $x(n)$ is related to a latent sharp RAW image $z(n)$ of the scene as
\begin{equation}
  \label{eq:dc_obs}
  x(n) = (k \circ z)(n) + \epsilon(n),
\end{equation}
where $k$ is the blur kernel and $\epsilon(n)$ is additive white Gaussian noise. The operator $\circ$ denotes convolution of the 3-channel image $z$ with a single-channel kernel $k$, implemented as the convolution of the kernel with each image channel separately. Although \eqref{eq:dc_obs} assumes convolution with a spatially-uniform blur kernel, the approach in this section can be easily generalized to account for non-uniform blur (\eg, as in~\cite{whyte2012non}). 

Deblurring an image involves estimating the blur kernel $k(n)$ acting on the image, and then inverting this blur to recover the sharp image $z(n)$. In this section, we will concentrate on this second step, \ie, deconvolution, assuming that the kernel $k$ has already been estimated--- say by applying the deterministic inverse and using a standard kernel estimation algorithm such as \cite{dblr0}\footnote{Empirically, we find that using a deterministic inverse suffices for the kernel estimation step, as it involves pooling information from the entire image to estimate a relatively small number of parameters.}. 

We begin with a modern linear-image deconvolution algorithm \cite{krishnan} and adapt it to exploit the inverse probability distributions from Sec.~\ref{sec:pinv}. Given an observed linear blurred image $x$ and known kernel $k$, Krishnan and Fergus~\cite{krishnan} provide a fast algorithm to estimate the latent sharp image $z(n)$ by minimizing the cost function
\begin{equation}
  \label{eq:dcost}
  \mathcal{C}(z) = \frac{\lambda}{2} \sum_n \| (k\circ z)(n) - x(n) \|^2 +  \sum_{n,i} \ \|(\nabla_i\circ z)(n)\|^\gamma,
\end{equation}
where $\{\nabla_i\}$ are gradient filters (horizontal and vertical finite difference filters in both \cite{krishnan} and our implementation), and the exponent $\gamma $ is $ \leq 1$. The first term measures the agreement of $z$ with the linear observation while the second term imposes a sparse prior on gradients in a sharp image. The scalar weight $\lambda$ controls the relative contribution of the two. 

Given the tone-mapped version $y(n)$ of the blurry linear image $x(n)$, the deterministic approach would be to simply replace $x(n)$ with its expected value $\mu(y(n))$ in the cost function above. However, to account for the structured uncertainty in our estimate of $x(n)$ and the fact that some values of $y(n)$ are more reliable than others, we modify the cost function to incorporate both the derendered means $\mu$ and co-variances $\Sigma$:
\begin{align}
  \label{eq:dcost2}
  \tilde{\mathcal{C}}(z) &= \frac{\lambda}{2} \sum_n \left[(k\circ z)(n) - \mu_n\right]^T\Sigma_n^{-1}\left[(k\circ z)(n) - \mu_n\right]^T\notag\\
&\qquad+ \sum_{n,i} \ \|(\nabla_i\circ z)(n)\|^\gamma,
\end{align}
where $\mu_n = \mu(y(n))$, $\Sigma_n = \Sigma(y(n)) + \sigma_z^2I_{3\times 3}$, and $\sigma_z^2$ is  
the expected variance of photo-sensor noise.

The algorithm in \cite{krishnan} minimizes the original cost function \eqref{eq:dcost} using an optimization strategy known as \emph{half-quadratic} splitting. It introduces a new set of auxiliary variables $w_i(n)$ corresponding to the gradients of the latent image $z(n)$, and carries out the following minimizations successively:
\begin{align}
  \label{eq:xsub}
  z = \arg \min_{z} &~\frac{\lambda}{2} \sum_n \| (k\circ z)(n) - x(n) \|^2\notag\\
  &~~ + \frac{\beta}{2} \sum_{n,i} \left\|(\nabla_i\circ z)(n)-w_i(n)\right\|^2,\\  
  w_i(n) = \arg \min_{w} &~\frac{\beta}{2} \sum_{n,i} \left\|(\nabla_i\circ z)(n)-w\right\|^2 + \|w\|^\gamma,
\end{align}
where $\beta$ is a scalar parameter that is increased across iterations. To minimize our modified cost function in \eqref{eq:dcost2}, we need only change \eqref{eq:xsub} to
\begin{align}
  \label{eq:xsubmod}
  z = \arg \min_{z} &~\frac{\lambda}{2} \sum_n \left[ (k\circ z)(n) - \mu_n\right]^T\Sigma_n^{-1}\left[ (k\circ z)(n) - \mu_n\right]\notag\\
  &~~ + \frac{\beta}{2} \sum_{n,i} \left\|(\nabla_i\circ z)(n)-w_i(n)\right\|^2.
\end{align}
A complication arises from this change: while the original expression \eqref{eq:xsub} can be computed in closed-form in the Fourier domain, the same is not true for the modified version \eqref{eq:xsubmod} because the first term has spatially-varying weights. Thus, to compute \eqref{eq:xsubmod} efficiently, we use a \emph{second} level of iterations based on variable-splitting to compute \eqref{eq:xsubmod} in every outer iteration of the original algorithm.

Specifically, we introduce a new cost-function with new auxiliary variables $u(n)$:
\begin{align}
  \label{eq:split2}
  z = \arg \min_{z} \min_{u} &~\frac{\lambda}{2} \sum_n \left[ u(n) - \mu_n\right]^T\Sigma_n^{-1}\left[ u(n) - \mu_n\right]\notag\\
  &~~ + \frac{\alpha}{2} \|(k\circ z)(n) - u(n) \|^2\notag\\
  &~~ + \frac{\beta}{2} \sum_{n,i} \left\|(\nabla_i\circ z)(n)-w_i(n)\right\|^2,
\end{align}
where $\alpha$ is a scalar variable whose value is increased across iterations. Note that for $\alpha \rightarrow \infty$, the expressions in \eqref{eq:xsubmod} and \eqref{eq:split2} are equivalent. The minimization algorithm proceeds as follows: we begin by setting $u(n) = \mu_n$ and then consider increasing values of $\alpha$ equally spaced in the $\log$ domain (in our implementation, we consider eight values that go from $4\lambda$ times the minimum to the maximum of all diagonal entries of all $\Sigma_n^{-1}$). For each value of $\alpha$, we perform the following updates to $z$ and $u(n)$ in sequence:
\begin{align}
  \label{eq:nupd1}
  z = \arg \min_z &~\frac{\alpha}{2} \|(k\circ z)(n) - u(n) \|^2\notag\\
  &~~ + \frac{\beta}{2} \sum_{n,i} \left\|(\nabla_i\circ z)(n)-w_i(n)\right\|^2,\\
  \label{eq:nupd2}
  u(n) = \arg \min_u &~\frac{\lambda}{2} \sum_n \left[ u - \mu_n\right]^T\Sigma_n^{-1}\left[ u - \mu_n\right]\notag\\
  &~~ + \frac{\alpha}{2} \|(k\circ z)(n) - u \|^2.
\end{align}
Note that \eqref{eq:nupd1} has the same form as the original \eqref{eq:xsub} and can be computed in closed-form in the Fourier domain. The updates to $u(n)$ in \eqref{eq:nupd2} can also be computed in closed-form independently for each pixel location $n$.

\begin{figure*}[!t]
  \centering
  \includegraphics[width=\textwidth]{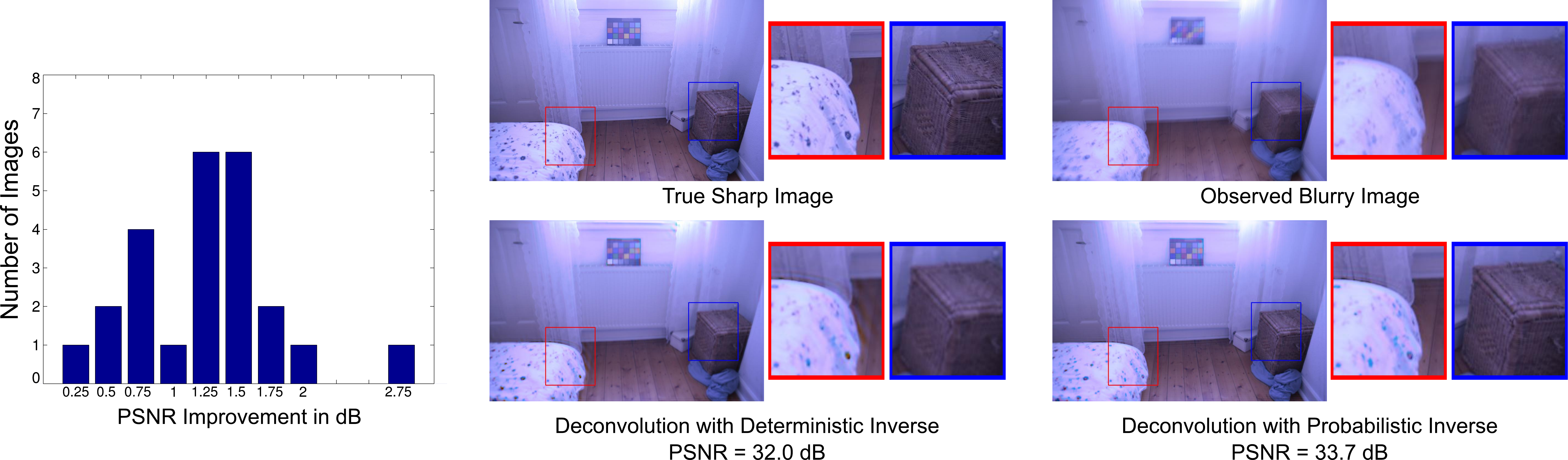}
  \caption{Deconvolution Results. (Left) Histogram of the improvement in PSNR across the twenty-four synthetically blurred images, when using a probabilistic approach instead of a deterministic one. (Right) Example deconvolution results using different approaches.}
  \label{fig:dcresults}
\end{figure*}

We evaluate the proposed algorithm using three RAW images from a public database~\cite{gehler,shidb} and eight (spatially-uniform) camera-shake blur kernels from the database in~\cite{dblr3}. We generate a set of twenty-four blurred JPEG observations by convolving each RAW image with each blur kernel, adding Gaussian noise, and then applying the estimated forward map of the Canon EOS-40D camera.\footnote{Note that for generating the synthetically blurred and ground-truth sharp images, we use the forward map as estimated using the uniformly sampled 8k training set, which as seen in Table~\ref{tab:fferrs} nearly perfectly predicts the camera map. During deconvolution, we use inverse and forward maps estimated using only the smaller ``8 exposures, 4 illuminants'' set.} We compare deconvolution performance of the proposed approach to a deterministic baseline consisting of the algorithm in \cite{krishnan} applied to the derendered means $\mu(n)$. The error is measured in terms of PSNR values between the true and estimated JPEG versions of the latent sharp image. Since these errors depend on the choice of regularization parameter $\lambda$ (which in practice is often set by hand), we perform a grid search to choose $\lambda$ separately for each of the deterministic and probabilistic approaches and for every observed image, selecting the value each time that gives the lowest RMSE using the known ground-truth sharp image. This measures the best performance possible with each approach. We set the exponent value $\gamma$ to $2/3$ as suggested in \cite{krishnan}.

Figure~\ref{fig:dcresults} shows a histogram of the improvement in PSNR across the different images when using the probabilistic approach over the deterministic one. The probabilistic approach leads to higher PSNR values for all images, with a median improvement of 1.24~dB. Figure~\ref{fig:dcresults} also includes an example of deconvolution results from the two approaches, and we see that in comparison to the probabilistic approach, the deterministic algorithm yields over-smoothed results in some regions while producing ringing artifacts in others. This is because the single scalar weight $\lambda$ is unable to adapt to the varying levels of ``noise'' or radiometric uncertainty in the derendered estimates. The ringing artifacts in the deterministic algorithm output correspond to regions of high uncertainty, where the probabilistic approach correctly employs a lower weight (\ie, $\Sigma_n^{-1}$) for the first term of \eqref{eq:dcost2} and smooths out the artifacts by relying more on the prior (\ie, the second term). At the same time, it yields sharper estimates in regions with more reliable observations by using a higher weight for the fidelity term, thus reducing the effect of the smoothness prior.

\section{Conclusion}
\label{sec:conc}

Traditionally, computer vision algorithms that require accurate linear measurements of spectral radiance have been limited to RAW input, and therefore to training and testing on small, specialized datasets. In this work, we present a framework that enables these methods to be extended to operate, as effectively as possible, on tone-mapped input instead. Our framework is based on incorporating a precise model of the uncertainty associated with global tone-mapping processes, and it makes it possible for computer vision systems to take better advantage of the vast number of tone-mapped images produced by consumer cameras and shared online.

To a vision engineer, our model of tone-mapping uncertainty is simply a form of signal-dependent Gaussian noise, and this makes it conceptually appealing for inclusion in subsequent visual processing. To prove this point, we introduced new, probabilistic adaptations of three classical inference tasks: image fusion, photometric stereo and deblurring. In all of these cases, an explicit characterization of the ambiguity due to tone-mapping allows the computer vision algorithm to surpass the performance possible with a purely deterministic approach.

In future work, the use of inverse RAW distributions should be incorporated in other vision algorithms, such as depth from stereo, structure from motion, and object recognition. This may require exploring approximations for the inverse distribution different from the Gaussian approximation in \eqref{eq:musdef}. While some applications might require more complex parametric forms, others may benefit from simpler weighting schemes that are derived based on the analysis in Sec.~\ref{sec:pinv}.

Also, it will be beneficial to find ways of improving calibration for JPEG-only cameras. Our hope is that eventually our framework will enable a common repository of tone-map models (or probabilistic inverse models) for each imaging mode of each camera, making the totality of Internet photos more usable by modern computer vision algorithms.

\section*{Acknowledgments}

The authors would like to thank the associate editor and reviewers for their comments. This material is based on work supported by the National Science Foundation under Grants no.~IIS-0905243, IIS-0905647, IIS-1134072, IIS-1212798, IIS-1212928, IIS-0413169, and IIS-1320715; by DARPA under the Mind's Eye and MSEE programs; and by Toyota.



\begin{thebibliography}{10}
\providecommand{\url}[1]{#1}
\csname url@samestyle\endcsname
\providecommand{\newblock}{\relax}
\providecommand{\bibinfo}[2]{#2}
\providecommand{\BIBentrySTDinterwordspacing}{\spaceskip=0pt\relax}
\providecommand{\BIBentryALTinterwordstretchfactor}{4}
\providecommand{\BIBentryALTinterwordspacing}{\spaceskip=\fontdimen2\font plus
\BIBentryALTinterwordstretchfactor\fontdimen3\font minus
  \fontdimen4\font\relax}
\providecommand{\BIBforeignlanguage}[2]{{%
\expandafter\ifx\csname l@#1\endcsname\relax
\typeout{** WARNING: IEEEtran.bst: No hyphenation pattern has been}%
\typeout{** loaded for the language `#1'. Using the pattern for}%
\typeout{** the default language instead.}%
\else
\language=\csname l@#1\endcsname
\fi
#2}}
\providecommand{\BIBdecl}{\relax}
\BIBdecl

\bibitem{mitsunaga1999rsc}
T.~Mitsunaga and S.~Nayar, ``Radiometric self calibration,'' in
  \emph{Proc.~CVPR}, 1999.

\bibitem{debevec97hdr}
P.~Debevec and J.~Malik, ``Recovering high dynamic range radiance maps from
  photographs,'' in \emph{SIGGRAPH}, 1997.

\bibitem{mann1995bud}
S.~Mann and R.~Picard, ``Being `undigital' with digital cameras: Extending
  dynamic range by combining differently exposed pictures,'' in \emph{Proc.
  IS\&T Annual Conf.}, 1995.

\bibitem{emor}
M.~D. Grossberg and S.~K. Nayar, ``Modeling the space of camera response
  functions,'' \emph{PAMI}, 2004.

\bibitem{bmvc09}
A.~Chakrabarti, D.~Scharstein, and T.~Zickler, ``An empirical camera model for
  internet color vision,'' in \emph{Proc. BMVC}, 2009.

\bibitem{kim2012new}
S.~Kim, H.~Lin, Z.~Lu, S.~S\"{u}sstrunk, S.~Lin, and M.~Brown, ``A new
  in-camera imaging model for color computer vision and its application,''
  \emph{PAMI}, 2012.

\bibitem{lee2013radiometric}
J.-Y. Lee, Y.~Matsushita, B.~Shi, I.~S. Kweon, and K.~Ikeuchi, ``Radiometric
  calibration by rank minimization,'' \emph{PAMI}, 2013.

\bibitem{photometric}
R.~Woodham, ``Photometric method for determining surface orientation from
  multiple images,'' \emph{Optical engineering}, 1980.

\bibitem{krishnan}
D.~Krishnan and R.~Fergus, ``Fast image deconvolution using {H}yper-{L}aplacian
  priors,'' in \emph{NIPS}, 2009.

\bibitem{lin2004rcs}
S.~Lin, J.~Gu, S.~Yamazaki, and H.-Y. Shum, ``Radiometric calibration from a
  single image,'' in \emph{Proc.~CVPR}, 2004.

\bibitem{TaiCKLYYMB:2013}
Y.~Tai, X.~Chen, S.~Kim, F.~Li, J.~Yang, J.~Yu, Y.~Matsushita, and M.~Brown,
  ``Nonlinear camera response functions and image deblurring: Theoretical
  analysis and practice,'' \emph{PAMI}, 2013.

\bibitem{farid2002blind}
H.~Farid, ``Blind inverse gamma correction,'' \emph{IEEE Trans.~Imag. Proc.},
  2002.

\bibitem{kuthirummal2008plp}
S.~Kuthirummal, A.~Agarwala, D.~Goldman, and S.~Nayar, ``Priors for large photo
  collections and what they reveal about cameras,'' in \emph{Proc.~ECCV}, 2008.

\bibitem{grossberg2003determining}
M.~Grossberg and S.~Nayar, ``Determining the camera response from images: what
  is knowable?'' \emph{PAMI}, 2003.

\bibitem{reinhard2006high}
E.~Reinhard, G.~Ward, S.~Pattanaik, and P.~Debevec, \emph{{High dynamic range
  imaging}}.\hskip 1em plus 0.5em minus 0.4em\relax Elsevier, 2006.

\bibitem{shi2010self}
B.~Shi, Y.~Matsushita, Y.~Wei, C.~Xu, and P.~Tan, ``{Self-calibrating
  photometric stereo},'' in \emph{Proc.~CVPR}, 2010.

\bibitem{ChenLYY:2012}
X.~Chen, F.~Li, J.~Yang, and J.~Yu, ``A theoretical analysis of camera response
  functions in image deblurring,'' \emph{Proc.~ECCV}, 2012.

\bibitem{pal2004probability}
C.~Pal, R.~Szeliski, M.~Uyttendaele, and N.~Jojic, ``Probability models for
  high dynamic range imaging,'' in \emph{Proc.~CVPR}, 2004.

\bibitem{GrundmannMKEK:2013}
M.~Grundmann, C.~McClanahan, S.~B. Kang, and I.~Essa, ``Post-processing
  approach for radiometric self-calibration of video,'' in \emph{Proc.~ICCP},
  2013.

\bibitem{xiong2012pixels}
Y.~Xiong, K.~Saenko, T.~Darrell, and T.~Zickler, ``From pixels to physics:
  {P}robabilistic color de-rendering,'' in \emph{Proc.~CVPR}, 2012.

\bibitem{libsvm}
C.-C. Chang and C.-J. Lin, ``{LIBSVM}: A library for support vector machines,''
  \emph{ACM Trans.~Intelligent Systems and Technology}, 2011, (software
  available at \url{http://www.csie.ntu.edu.tw/~cjlin/libsvm}).

\bibitem{holm2006capture}
J.~Holm, ``Capture color analysis gamuts,'' in \emph{IS\&T/SID Color and
  Imaging Conference}, 2006.

\bibitem{jiang2013space}
J.~Jiang, D.~Liu, J.~Gu, and S.~S{\"u}sstrunk, ``What is the space of spectral
  sensitivity functions for digital color cameras?'' in \emph{IEEE Workshop on
  the Applications of Computer Vision (WACV)}, 2013.

\bibitem{projectpage}
\BIBentryALTinterwordspacing
 [Online]. Available: \url{http://vision.seas.harvard.edu/derender/}
\BIBentrySTDinterwordspacing

\bibitem{shen2009photometric}
L.~Shen and P.~Tan, ``Photometric stereo and weather estimation using internet
  images,'' in \emph{Proc.~CVPR}, 2009.

\bibitem{lalonde2010sun}
J.~Lalonde, S.~Narasimhan, and A.~Efros, ``What do the sun and the sky tell us
  about the camera?'' \emph{IJCV}, 2010.

\bibitem{ackermann2012photometric}
J.~Ackermann, F.~Langguth, S.~Fuhrmann, and M.~Goesele, ``Photometric stereo
  for outdoor webcams,'' in \emph{Proc.~CVPR}, 2012.

\bibitem{barsky2001colour}
S.~Barsky and M.~Petrou, ``Colour photometric stereo: Simultaneous
  reconstruction of local gradient and colour of rough textured surfaces,'' in
  \emph{Proc.~ICCV}, 2001.

\bibitem{elementOfStatisticalLearning}
T.~Hastie, R.~Tibshirani, and J.~Friedman, \emph{The Elements of Statistical
  Learning: Data Mining, Inference, and Prediction}.\hskip 1em plus 0.5em minus
  0.4em\relax Springer, 2009.

\bibitem{frankotIntegration}
R.~Frankot and R.~Chellappa, ``A method for enforcing integrability in shape
  from shading algorithms,'' \emph{PAMI}, 1988.

\bibitem{dblr0}
R.~Fergus, B.~Singh, A.~Hertzmann, S.~Roweis, and W.~Freeman, ``{Removing
  camera shake from a single photograph},'' in \emph{SIGGRAPH}, 2006.

\bibitem{dblr1}
Q.~Shan, J.~Jia, and A.~Agarwala, ``{High-quality motion deblurring from a
  single image},'' in \emph{SIGGRAPH}, 2008.
\newpage
\bibitem{dblr2}
J.~Cai, H.~Ji, C.~Liu, and Z.~Shen, ``{Blind motion deblurring from a single
  image using sparse approximation},'' in \emph{Proc.~CVPR}, 2009.

\bibitem{dblr3}
A.~Levin, Y.~Weiss, F.~Durand, and W.~Freeman, ``{Understanding and evaluating
  blind deconvolution algorithms},'' in \emph{Proc.~CVPR}, 2009.

\bibitem{whyte2012non}
O.~Whyte, J.~Sivic, A.~Zisserman, and J.~Ponce, ``Non-uniform deblurring for
  shaken images,'' \emph{IJCV}, 2012.

\bibitem{kimdb}
S.~Kim, Y.-W. Tai, S.~J. Kim, M.~S. Brown, and Y.~Matsushita, ``Nonlinear
  camera response functions and image deblurring,'' in \emph{Proc.~CVPR}.\hskip
  1em plus 0.5em minus 0.4em\relax IEEE, 2012.

\bibitem{gehler}
P.~V. Gehler, C.~Rother, A.~Blake, T.~Minka, and T.~Sharp, ``Bayesian color
  constancy revisited,'' in \emph{Proc.~CVPR}, 2008.

\bibitem{shidb}
L.~Shi and B.~Funt, ``Re-processed version of the {G}ehler color constancy
  dataset of 568 images,'' 2010, accessed from
  \url{http://www.cs.sfu.ca/~colour/data/}.

\end{thebibliography}
\end{document}